\documentclass{article}

\usepackage{arxiv}
\usepackage{natbib}
\renewcommand{\headeright}{}
\usepackage[utf8]{inputenc} 
\usepackage[T1]{fontenc}    
\usepackage{hyperref}       
\usepackage{url}           
\usepackage{booktabs}       
\usepackage{nicefrac}       
\usepackage{microtype}   
\usepackage{lipsum}        
\usepackage{doi}
\usepackage{lineno}
\usepackage{soul}
\usepackage{amsmath,amssymb,amsfonts,amsthm}
\usepackage{mathtools}
\usepackage{float}
\usepackage{cancel}
\usepackage{placeins}
\usepackage{array}
\usepackage{tabularx}
\usepackage{multirow}
\usepackage{multicol}
\usepackage{makecell}
\usepackage{cases}
\usepackage[linesnumbered,ruled]{algorithm2e}
\usepackage{graphicx}
\graphicspath{{figure/}}
\usepackage{subcaption}
\usepackage{cleveref} 

\newcommand{\vect}[1]{\boldsymbol{\mathrm{#1}}}

\newcommand{\orcidicon}{%
    \includegraphics[scale=0.06]{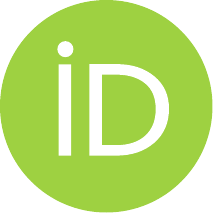}%
}

\newcolumntype{L}[1]{>{\raggedright\arraybackslash}m{#1}}
\newcolumntype{C}[1]{>{\centering\arraybackslash}m{#1}}
\newcolumntype{R}[1]{>{\raggedleft\arraybackslash}m{#1}}

\title{Approximate Adjoint Models for Memory and Compute-Efficient Online Learning of Hybrid Models}

\date{}

\newif\ifuniqueAffiliation
\uniqueAffiliationtrue
\ifuniqueAffiliation 

\author{
    \href{https://orcid.org/0009-0000-2141-7153}{\orcidicon\hspace{1mm}Emilio González-Zamora}$^{1,2}$
    \thanks{Corresponding author: \texttt{emilio.gonzalez-zamora@imt-atlantique.fr}},
    \href{https://orcid.org/0000-0003-1647-8239}{\orcidicon\hspace{1mm}Pierre Tandeo}$^{1,2}$,
    \href{https://orcid.org/0009-0005-2419-7949}{\orcidicon\hspace{1mm}Laurent Debreu}$^{3}$,
    \href{https://orcid.org/0000-0003-0554-3971}{\orcidicon\hspace{1mm}Said Ouala}$^{1,2}$
    \\[1em]
    \small
    $^1$ IMT Atlantique, Lab-STICC, UMR CNRS 6285, 29238, Brest, France
    \\
    $^2$ ODYSSEY Team-Project, INRIA IMT-Atlantique CNRS, Brest, France
    \\
    $^3$ Univ. Grenoble Alpes, Inria, CNRS, Grenoble INP, LJK, Grenoble, France
}

\renewcommand{\shorttitle}{}

\hypersetup{
    pdftitle={Approximate Adjoint Models for Memory and Compute-Efficient Online Learning of Hybrid Models},
    pdfauthor={Emilio González-Zamora, Pierre Tandeo, Laurent Debreu, Said Ouala},
    pdfsubject={Approximate adjoint methods and online learning of hybrid numerical models},
    pdfkeywords={Differentiable programming, hybrid models, online learning, machine-Learning, numerical simulation, gradient approximation, Optimization}
}

\begin{document}
\maketitle
\begin{abstract}
	Differentiable programming enables online calibration of numerical models by minimizing errors accumulated along simulated trajectories. This includes estimating physical parameters and training neural sub-models in hybrid configurations. Gradients are typically computed using adjoint methods, or backpropagation through time (BPTT), whose computational and memory costs grow with the length of the training window. This limits the duration and resolution of hybrid-model training, a major bottleneck for realistic geophysical applications. Approaches like truncated BPTT reduces memory use by discarding sensitivities across truncation boundaries, introducing gradient bias and neglecting long-range dependencies. We propose a class of gradient approximation methods that preserve the adjoint recurrence over the full trajectory while approximating only the pullback operators within it. Exact pullbacks are computed by automatic differentiation at sparse anchor states only. Between anchors, sensitivities propagate either through the identity (Zero-order) or through a reused, frozen pullback (Pullback Reuse). This approximation makes the parameter pullback constant within each chunk, allowing the gradient sum to factorize so that a single evaluation per chunk replaces one per time step. We evaluate both on three systems of increasing complexity, an incomplete Lorenz-63 model, a large-eddy closure for two-dimensional quasi-geostrophic turbulence, and bias correction in a multilayer quasi-geostrophic ocean model. Peak memory falls by up to an order of magnitude and runtime up to fourfold, with Pullback Reuse staying closest to the exact gradient direction. In all three cases, approximate gradients support hybrid-model training at comparable accuracy, showing that effective online learning does not require exact gradients.
\end{abstract}

\keywords{Differentiable programming \and Hybrid models \and Online learning \and Machine-Learning \and Numerical simulation \and Gradient approximation \ Optimization}

\section{Introduction}\label{sec:intro}
Numerical simulations of fluid flows require representing the interactions between processes that span a continuum of scales. Direct Numerical Simulations (DNS) resolve them explicitly, on a grid fine enough to capture the smallest dissipative scale, which is tractable only at moderate Reynolds numbers and in  idealized configurations \citep{moin_and_mahesh_1998_dns, pope_2000_turbulent_flows, sandham_etal_2017_dns_reynolds}. For realistic flows, Computational Fluid Dynamics and its geophysical counterparts instead seek the best fidelity attainable under a fixed computational budget. Common strategies rely on decomposing the flow into a numerically resolved large-scale and an unresolved remainder whose effect on the resolved scale is represented by a sub-model or closure. Reynolds-Averaged Navier-Stokes (RANS) and Large Eddy Simulations (LES) are canonical examples \citep{leonard_1975_energy_les,mason_1994_les_review, meneveau_and_katz_2000_turbulence_closure_les, ghosal_1999_les_turbulence, pope_2000_turbulent_flows}. 

Physics-based sub-models derive this closure from first principles, as in eddy-viscosity assumptions where the unresolved scales are treated as mainly diffusive \citep{smagorinsky_1963_general_circulation, germano_etal_1991_dynamic_subgrid_eddy_viscosity, spalart_and_almarras_1992_one_eq_tubulence_model, wilcox_2008_komega_turbulence_model}; similar heuristics are used for unresolved processes in the ocean and atmosphere \citep{christensen_and_zanna_2022_parametrization_weather_climate_models}. Their continuous form permits theoretical validation and analysis of the well-posedness of the resulting equations, but they remain a significant source of uncertainty. Their free parameters are difficult to tune, and assumptions they encode do not always hold in the regimes where they are applied. Unresolved scales, for instance, do not only dissipate energy but they redistribute it and feed it back to larger scales, which diffusive closures cannot represent \citep{piomelli_etal_1991_sgs_backscatter, kitsios_etal_2012_sgs_param_ocean_scaling_laws, jansen_and_held_2014_sgs_energy_backscatter, zanna_etal_2017_stochastic_param_eddy_mean_flow}. 

The increasing availability of high-resolution simulations and observations, together with the progress in deep learning, has motivated a growing interest in the use of machine learning and artificial intelligence for numerical modeling. A first line of work replaces the numerical model entirely by a data-driven surrogate, either through reduced-order models evolving a low-dimensional representation of the flow \citep{ouala_etal_2020_learning_latent_dynamics,ouala_etal_2023_latent_edmd,ouala_etal_2023_bounded}, or through neural emulators trained to reproduce the dynamics of the full state \citep{pathak_etal_2022_fourcastnet, bi_etal_2023_pangu_weather, lam_etal_2023_graphcast}. A second approach, known as hybrid modeling, retains the physical dynamical model and adds a learned component to account for unresolved dynamics, correct systematic model biases, or replace uncertain closure schemes. This learned component is typically a deep neural network \citep{rasp_2020_coupled_online,ouala_etal_2024_online_calib_submodels,frezat_etal_2022_aposteriori_qg_turbulence_param,maddison_2026_online_learning_gyres}.

Two strategies exist for training  the learned component of a hybrid model. Offline or \emph{a priori} learning, trains the sub-model independently of the dynamical system, as a supervised regression problem against a reference response. This approach remains widely adopted for its conceptual simplicity and low computational cost \citep{zanna_2021_DL_turbulence,frezat_etal_2021_physical_invariance_nn}.Because the sub-model never interacts with the physics during training, however, it is sensitive to the distribution shift between the training data and the states encountered during inference, and offline-trained parameterizations may show degraded long-term predictive skill, unphysical behavior, or numerical instabilities once coupled to the dynamical model \citep{ frezat_etal_2022_aposteriori_qg_turbulence_param,maddison_2026_online_learning_gyres, yan_etal_2025_adjoint_online_qg}. 

Online or \emph{a posteriori} learning, addresses these limitations by embedding the sub-model within the dynamical system during training. The coupled hybrid system is unrolled over several time steps, the loss is evaluated on the resulting trajectory, and the gradients are propagated through both the dynamical model and the sub-model. This formulation typically relies on differentiable programming where the numerical solver is written in a framework that supports automatic differentiation, so that the sensitivity of the simulated trajectory to the sub-model parameters is available. The sub-model therefore learns on the states the coupled system actually visits, and accounts for the temporal dependencies between them by minimizing the error accumulated along the unrolled hybrid trajectory. Online learning has been shown to improve forecast skill, promote long-term stability, and reduce the volume of training data required \citep{ frezat_etal_2022_aposteriori_qg_turbulence_param, list_etal_2022_learned_turbulence_diff, ouala_etal_2024_online_calib_submodels}. 

Beyond idealized cases, current work aims to demonstrate online learning on realistic, large-scale problems, with recent examples in atmosphere \citep{kochkov_neural_2024} and ocean \citep{meunier_etal_2025_veros} modeling. However, training through long simulated trajectories remains computationally expensive and thus the major bottleneck. Exact gradients requires information from intermediate simulation steps to be available during the backward pass, and thus it requires storing every intermediate states before  traversing them backwards, making memory grow with both the trajectory length and the dimension of the model state. For systems with many degrees of freedom, this can limit the duration of the training window and the resolution at which the hybrid model can be trained. This constraint may help explain why some previous studies have relied on relatively coarse resolutions and limited domains  \citep{kochkov_neural_2024, meunier_etal_2025_veros}. Approaches such as checkpointing \citep{griewank_and_walther_2000_checkpointing_revolve, stumm_and_walther_2010_online_checkpointing, chen_etal_2016_sqrt_checkpointing} and truncated backpropagation through time (TBPTT) offer different ways to manage these costs, but involve trade-offs between memory, computation, and gradient accuracy. We discuss these approaches in more detail in Section \ref{sec:related_works}.
 
In this work, we target the computational cost of online learning. We introduce a class of Pullback Approximation methods that reduce the cost of the backward pass while propagating sensitivities across the full trajectory. The paper is organized as follows. Section \ref{sec:problem_statement} formalizes the hybrid modeling framework and the online learning problem. Section \ref{sec:grad_approx} introduces the proposed gradient approximation methods within a unified framework that also covers backpropagation through time and its truncated variant, Truncated Backpropagation Through Time (TBPTT), which serves as our biased-gradient baseline. Section \ref{sec:related_works} situates them with respect to other strategies for reducing the cost of exact gradients, and to related work in numerical analysis and deep learning. Section \ref{sec:experimental_setup_results} describes the case studies and reports the results, evaluating methods in terms of both model accuracy and computational cost. Section \ref{sec:discussion_conclusion} discusses the implications and outlines directions for future work.

\section{Problem statement}\label{sec:problem_statement} 

Let us consider a hybrid dynamical system governed by the following parameterized ordinary differential equation (ODE):
\begin{equation}
    \dot{\vect{u}}_t = \mathbf{F}(\mathbf{u}_t) +
    \mathbf{M}(\mathbf{u}_t,\vect{\theta})
    \label{eq:ode_system}
\end{equation}
where $\vect{u}_t$ $\in$ $\mathbb{R}^d$ denotes the state of the system at time $t$. The right-hand side is composed of two contributions. The first term, $\vect{F}$, represents the physics-based dynamical model which is derived from first principles or established empirical formulations. The second term, $\mathbf{M}$, is a data-driven sub-model designed to account for imperfections or unmodeled dynamics in $\vect{F}$. This component is typically implemented as a neural network parameterized by a set of trainable parameters $\vect{\theta}\in\mathbb{R}^a$.

In practice, the continuous-time system in Eq.\eqref{eq:ode_system} is discretized and evolved in time using a numerical time-integration scheme. We therefore introduce a mapping $f$ that represents the numerical integrator of the system \eqref{eq:ode_system}. Given an initial state $\vect{u}_0 \in \mathbb{R}^d$, and a parameter vector, the system evolves through the recursive application of $f$ to generate a rollout trajectory $\vect{u}_{n} = f(\vect{u}_{n-1}, \vect{\theta})$, where  $n\in\mathbb{Z}^{+}$ denotes the discrete time index. 

In online learning, the optimization problem is formulated directly on such trajectories and the objective is to identify $\vect{\theta}$ that minimizes a loss function evaluated over a specified finite temporally unrolled window. Let $N$ denote the number of integration steps over which the hybrid system is recursively unrolled. The state reached after this rollout can be expressed as
\begin{equation}
\vect{u}_N
=
\underbrace{
f \circ f \circ \cdots \circ f
}_{N\ \text{times}}
(\vect{u}_0),
\label{eq}
\end{equation}
In the online learning literature, $N$ is commonly referred to as the rollout length or unrolled window \citep{um_solver_in_loop,frezat_etal_2022_aposteriori_qg_turbulence_param,ouala_etal_2024_online_calib_submodels,list_etal_2025_unrolled_training}. The learning problem can thus be formulated as the minimization of a global loss function $\mathcal{L}_N$, which depends on the sequence of hybrid states $\left\{ \vect{u}_n\right\}_{n=1}^{N}$
\begin{equation}
    \underset{\vect{\theta}}{\text{min}} \quad \mathcal{L}_N(\left\{ \vect{u}_n = f(\vect{u}_{n-1},\vect{\theta})\right\}_{n=1}^{N}, \left\{ \vect{y}_n\right\}_{n=1}^{N},\vect{\theta})
    \label{eq:optimization_global}
\end{equation}
where $\mathcal{L}_N$ measures the discrepancy between the predicted trajectory and a target trajectory $\left\{ \vect{y}_n\right\}_{n=1}^{N}$.

As in any gradient-descent based optimization procedure, training requires computing the sensitivity of the loss with respect to the trainable parameters $\frac{d\mathcal{L}_N}{d\vect{\theta}}$. This quantity measures how perturbations of the sub-model parameters influence the trajectory-level objective through the coupled hybrid dynamics. In practice, these gradients are typically obtained using Backpropagation Through Time (BPTT), which is reviewed in the following section.

\section{Gradient Approximation Methods for Backpropagation Through Time}\label{sec:grad_approx}

\subsection{Backpropagation Through Time}\label{subsec:bptt}

In BPTT, the total gradient of the loss with respect to the parameters is expressed as in \citet{tallec_&_yann_2017_unbiased_bptt}:
\begin{equation}
\frac{d \mathcal{L}_N}{d \vect{\theta}} = \sum_{n=1}^{N} \vect{g}_n \frac{\partial f}{\partial \vect{\theta}}(\vect{u}_{n-1}, \vect{\theta})
\label{eq:bptt_gradient}
\end{equation}
where $\vect{g}_n = \frac{d \mathcal{L}_N}{d u_{n}}$ is the adjoint vector at step $n$, that is the sensitivity of the scalar loss with respect to the intermediate state $\vect{u}_n$. By the chain rule, the sequence of adjoint vectors satisfy an inhomogeneous backward recurrence:
\begin{subequations}
\begin{numcases}{}
    \vect{g}_N = \frac{d \mathcal{L}_N}{d \vect{u}_{N}} = \frac{\partial \mathcal{L}_N}{\partial \vect{u}_N}
    & \label{eq:adjoint_recurrence_a}\\
    \vect{g}_n = \frac{d \mathcal{L}_N}{d \vect{u}_{n}} = \vect{g}_{n+1} \frac{\partial f}{\partial \vect{u}}(\vect{u}_n, \vect{\theta}) + \frac{\partial \mathcal{L}_N}{\partial \vect{u}_n}
    & \label{eq:adjoint_recurrence_b}
\end{numcases}
\label{eq:adjoint_recurrence}
\end{subequations}
In this recurrence, the vector-matrix product $\vect{g}_{n+1} \frac{\partial f}{\partial \vect{u}}$ is a pullback operation along $f$, where the adjoint vector, a cotangent on the state space  $\vect{g}_{n+1}$ is pulled back from step $n+1$ to step $n$ by contraction with the Jacobian matrix of $f$ evaluated at $\vect{u}_n$. The additional source term $\frac{\partial \mathcal{L}_N}{\partial \vect{u}_n}$ accounts for the direct dependence of the loss on a given state, independently of the adjoints propagated from later steps.

In differentiable programming, a pullback is constructed by reverse-mode automatic differentiation (AD) at a given linearization state, and each of its evaluations is a vector-Jacobian product (VJP). Two distinct pullbacks appear in the computation. The state-pullback contracts $g_{n+1}$ with $\partial f/\partial \vect{u}$ to propagate the adjoint backward through \eqref{eq:adjoint_recurrence}, while the parameter-pullback contracts $g_n$ with $\partial f/\partial \theta$ to accumulate the parameter gradient in \eqref{eq:bptt_gradient}. We write $\text{VJP}^{\vect{u}}$ and $\text{VJP}^{\vect{\theta}}$ for their respective evaluations. 

Because the recurrence in \eqref{eq:adjoint_recurrence} is traversed backward in time, while each pullback is linearized at the corresponding forward state, evaluating it in general requires storing the entire sequence of intermediate forward states, together with the associated reverse-mode information, often referred to as the computation graph or tape. As the rollout length N increases, the memory required to store this graph grows linearly, quickly becoming a prohibitive bottleneck \citep{tallec_&_yann_2017_unbiased_bptt, sapienza_etal_2025_diff_progra}. 

\subsection{Truncated Backpropagation Through Time}\label{subsec:tbptt}

Truncated Backpropagation Through Time (TBPTT) reduces the computational and memory cost of BPTT by partitioning the unrolled trajectory into independent temporal windows, where the standard adjoint recurrence \eqref{eq:adjoint_recurrence} is retained only within each window. The memory required for the backward pass is thereby reduced, since the tape then spans a single window rather that the full trajectory. This method was first introduced for training Recurrent Neural Networks (RNNs) \citep{william_and_zipser_1995_tbptt} and was later explored in data assimilation frameworks and hybrid modeling frameworks \citep{list_etal_2025_unrolled_training}. For completeness, and to allow a comparison with the methods proposed below, we describe TBPTT in what follows.

We parameterize the degree of truncation by the window length $A_{\mathrm{TB}}$, expressed in unrolled time steps. For a trajectory of $N$ steps, the corresponding number of windows is $K$ such that $A_{\text{TB}} = N/{K}$. Applying the truncation to \eqref{eq:adjoint_recurrence} yields
\begin{subequations}
  \begin{numcases}{\vect{g}_n =}
        \frac{\partial \mathcal{L}_N}{\partial \vect{u}_n} & if $n \equiv 0 \pmod{A_{\text{TB}}}$, \label{eq:tbptt_recurrence_a}\\
        \mathbf{g}_{n+1} \frac{\partial f}{\partial \vect{u}}(\vect{u}_{n-1} , \vect{\theta}) + \frac{\partial \mathcal{L}_N}{\partial \vect{u}_n}& otherwise. \label{eq:tbptt_recurrence_b}
    \end{numcases}
    \label{eq:tbptt_recurrence}
\end{subequations}
At  a truncation boundary the incoming adjoint is discarded and only the local source term is retained, so that no sensitivity crosses from one window to the next. 

TBPTT therefore introduces a trade-off between memory usage and gradient fidelity. Increasing $A_{\mathrm{TB}}$ widens the backward window and retains richer long-range dependencies, at the cost of higher peak memory. Truncation changes neither the number of pullbacks constructed nor the number of VJPs evaluated over the trajectory; it only reduces how much of the tape must be held at once. However, discarding the incoming sensitivities across truncation boundaries introduces a systematic bias in the resulting gradient \citep{tallec_&_yann_2017_unbiased_bptt, 
tang_2018_training_rnn_tbptt, aicher_2020_adaptive_bptt,list_etal_2025_unrolled_training, schnell_2024_stabilizing_bptt_complex_physics}.

\subsection{Pullback Approximation Methods for Backpropagation Through Time}\label{subsec:pullba_approx_grad}

Rather than discarding sensitivity information at window boundaries as TBPTT does, we propose a new class of gradient approximation methods that preserve the adjoint recurrence \eqref{eq:adjoint_recurrence} over the full trajectory, while locally approximating the pullback operators within instead. The resulting gradients remain biased, but retain the long-range dependency structure of the exact adjoint. We refer to these as Pullback Approximation (PA) methods. 

Similarly to TBPTT, PA partitions the trajectory of $N$ steps into $K$ chunks
$C_1, \ldots, C_K$ of length $A_{PA} = N/K$. Each chunk $C_i$ has a single
anchor step $n_i^{\text{anc}}$ at its midpoint, whose corresponding state is
the linearization point for the whole chunk; exact state and parameter
pullbacks are constructed by reverse-mode AD at $n_i^{\text{anc}}$ only, and
reused or replaced at every other step of $C_i$. Under this approximation,
the parameter gradient \eqref{eq:adjoint_recurrence} thus becomes
\begin{equation}
\frac{d \mathcal{L}_N}{d \vect{\theta}}
\;\approx\;
\sum_{n=1}^{N} \vect{g}^{m}_n \,
\frac{\partial f}{\partial \vect{\theta}}
\big(\vect{u}_{n_i^{anc}}, \vect{\theta}\big)
\label{eq:pa_gradient}
\end{equation}
so that the parameter pullback at step n is linearized at the anchor state of its chunk rather than at $\vect{u}_{n-1}$. The approximate adjoint $g^m_n$ is obtained from a modified form of the recurrence \eqref{eq:adjoint_recurrence}, with $m$ indexing the two PA methods introduced below, \textit{Zero-order} (ZO) and \textit{Pullback Reuse} (Reuse).

\subsubsection{Zero-order approximation}
\label{subsec:zero_order}

The Zero-order (ZO) approximation replaces the state pullback at non-anchor steps by the identity. This follows from the Taylor expansion of the time-stepping function $f$ with step size $h$,
\begin{equation}
    \frac{\partial f}{\partial \vect{u}}(\vect{u}_n, \vect{\theta}) = \mathbf{I} +
    \sum_{k=1}^{p} \frac{h^k}{k!} \frac{\partial}{\partial \vect{u}}
    \frac{d^{k-1}(\mathbf{F}(\cdot) +
    \mathbf{M}(\cdot,\vect{\theta}))(\vect{u}_t)}{dt^{k-1}}\bigg|_{\vect{u}_n} + \mathcal{O}(h^{p+1}).
    \label{eq:exact_jacobian}
\end{equation}

in which every term beyond the leading one is $\mathcal{O}(h)$, so the Jacobian can be approximated as 
\begin{equation}
    \frac{\partial f}{\partial \vect{u}}(\vect{u}_n, \vect{\theta}) = \mathbf{I} + \mathcal{O}(h)
    \label{eq:zero_order_approx}
\end{equation}
Assuming that $h$ is small enough, the adjoint is then propagated unchanged between anchors, and the recurrence \eqref{eq:adjoint_recurrence}becomes, for every $n \in C_i$, 
\begin{subequations}
    \begin{numcases}
    {\vect{g}_n^{\text{ZO}} =}
       \vect{g}^{\text{ZO}}_{n+1} \frac{\partial f}{\partial \vect{u}}(\vect{u}_n, \vect{\theta}) + \frac{\partial \mathcal{L}_N}{\partial \vect{u}_n} & \text{if $n = n_i^{\text{anc}}$}, \\
    \vect{g}^{\text{ZO}}_{n+1} + \frac{\partial \mathcal{L}_N}{\partial \vect{u}_n} & \text{otherwise},
    \end{numcases}
    \label{eq:zero_order_recurrence}
\end{subequations}

\subsubsection{Pullback Reuse}
\label{subsec:reuse}

By replacing the state pullback with the identity, ZO discards the local dynamical information carried by the intermediate Jacobians. Pullback Reuse (Reuse) retains it at no additional memory cost, by reusing both, the anchor state and parameter pullbacks across the whole chunk rather than reconstructing them at every step, a strategy long used in implicit nonlinear solvers, discussed in\ref{sec:related_works}. This yields, for every $n \in C_i$,

\begin{subequations}
    \begin{numcases}
    {\vect{g}_n^{\text{Reuse}} =}
      \vect{g}^{\text{Reuse}}_{n+1} \frac{\partial f}{\partial \vect{u}}(\vect{u}_n, \vect{\theta}) + \frac{\partial \mathcal{L}_N}{\partial \vect{u}_n} & \text{if $n = n_i^{\text{anc}}$}, \\
      \vect{g}^{\text{Reuse}}_{n+1} \frac{\partial f}{\partial \vect{u}}(\vect{u}_{n_i^{\text{anc}}}, \vect{\theta}) + \frac{\partial \mathcal{L}_N}{\partial \vect{u}_n} & \text{otherwise},
    \end{numcases}
    \label{eq:reuse_recurrence}
\end{subequations}
The resulting approximation error scales with the variation of the local Jacobian over the chunk, and vanishes as $h \to 0$ or $A_{\text{PA}} \to 1$.

\subsection{Numerical implementation of the Pullback Approximation methods}\label{subsec:num_algo}

The computational efficiency of both PA strategies stems from a factorization of the parameter gradient accumulation. To see this, it is useful to write the global sum \eqref{eq:pa_gradient} chunk by chunk. Since the linearization point $\vect{u}_{\mathrm{a}(n)}$ is the same for all steps belonging to a given chunk, the sum splits into $K$ groups of $A_{\text{PA}}$ consecutive terms:

\begin{equation}
\begin{aligned}
\frac{d \mathcal{L}_N}{d \vect{\theta}} & \approx 
\sum_{n=1}^{N} \vect{g}^{m}_n \,
\frac{\partial f}{\partial \vect{\theta}}
\big(\vect{u}_{n_i^{anc}}, \vect{\theta}\big) \\
&=
\sum_{n=1}^{A_{\text{PA}}} \vect{g}^{m}_n \,
\frac{\partial f}{\partial \vect{\theta}}
\big(\vect{u}_{n_1^{\text{anc}}}, \vect{\theta}\big) +  \sum_{n=A_{\text{PA}}+1}^{2A_{\text{PA}}} \vect{g}^{m}_n \,
\frac{\partial f}{\partial \vect{\theta}}
\big(\vect{u}_{n_2^{\text{anc}}}, \vect{\theta}\big) \\
& +\ \cdots + \sum_{n=(K-1)A_{\text{PA}}+1}^{N} \vect{g}^{m}_n \, 
\frac{\partial f}{\partial \vect{\theta}}
\big(\vect{u}_{n_K^{\text{anc}}}, \vect{\theta}\big).
\end{aligned}
\label{eq:pa_gradient_split}
\end{equation}
where $\vect{g}_n^m$ follows either ZO or Reuse recurrence given in \eqref{eq:zero_order_recurrence} and \eqref{eq:reuse_recurrence} respectively.

Within each group, the parameter Jacobian no longer depends on the summation index $n$. By linearity of the vector-Jacobian product, it can therefore be pulled out of the sum, so that the $i$-th group reduces to a single product:
\begin{equation}
\sum_{n=(i-1)A_{\text{PA}}+1}^{i A_{\text{PA}}}
\vect{g}^{m}_n \,
\frac{\partial f}{\partial \vect{\theta}}
\big(\vect{u}_{n_i^{\text{anc}}}, \vect{\theta}\big)
=
\underbrace{\left( \sum_{n=(i-1)A_{\text{PA}}+1}^{i A_{\text{PA}}} \vect{g}^{m}_n \right)}_{\vect{S}_i}
\frac{\partial f}{\partial \vect{\theta}}
\big(\vect{u}_{n_i^{\text{anc}}}, \vect{\theta}\big),
\label{eq:chunk_factorization}
\end{equation}
where $\vect{S}_i$ is simply the sum of the approximate adjoints visited inside chunk $C_i$. The parameter gradient then reads

\begin{equation}
\frac{d \mathcal{L}_N}{d \vect{\theta}}
\;\approx\;
\sum_{i=1}^{K} \vect{S}_i \,
\frac{\partial f}{\partial \vect{\theta}}
\big(\vect{u}_{n_i^{\text{anc}}}, \vect{\theta}\big).
\label{eq:pa_gradient_factorized}
\end{equation}

Eq.\eqref{eq:pa_gradient_factorized} translates into an algorithm in which the backward pass follows a double-loop structure: an inner loop that accumulates the adjoint sensitivities within a chunk, and an outer loop that constructs the anchor pullbacks and evaluates \eqref{eq:pa_gradient_factorized}. It can be summarized as follows:
\begin{enumerate}
\item \textit{Outer loop over chunks (anchor linearization and parameter pullback).} The chunks are traversed backward, $i = K, \dots, 1$. Reverse-mode AD is invoked once at the anchor state $\vect{u}_{n_i^{\text{anc}}}$ to construct the two pullback operators $\text{VJP}_i^{\vect{u}}$ and $\text{VJP}_i^{\vect{\theta}}$. Once the inner loop has completed, $\text{VJP}_i^{\vect{\theta}}(\vect{S}_i)$ is added to the global parameter gradient.
\item \textit{Inner loop over steps (adjoint sensitivity accumulation).} The steps of the chunk are traversed backward. At each step, the loss cotangent $\frac{\partial \mathcal{L}_N}{\partial \vect{u}_n}$ is incorporated, the state sensitivity is updated according to the chosen approximation, the identity for ZO or the reused operator $\text{VJP}_i^{\vect{u}}$ for Reuse; the resulting $\vect{g}^{m}_n$ is accumulated into $\vect{S}_i$.
\end{enumerate}

\begin{figure}[ht]
    \centering
    \includegraphics[width=1\linewidth]{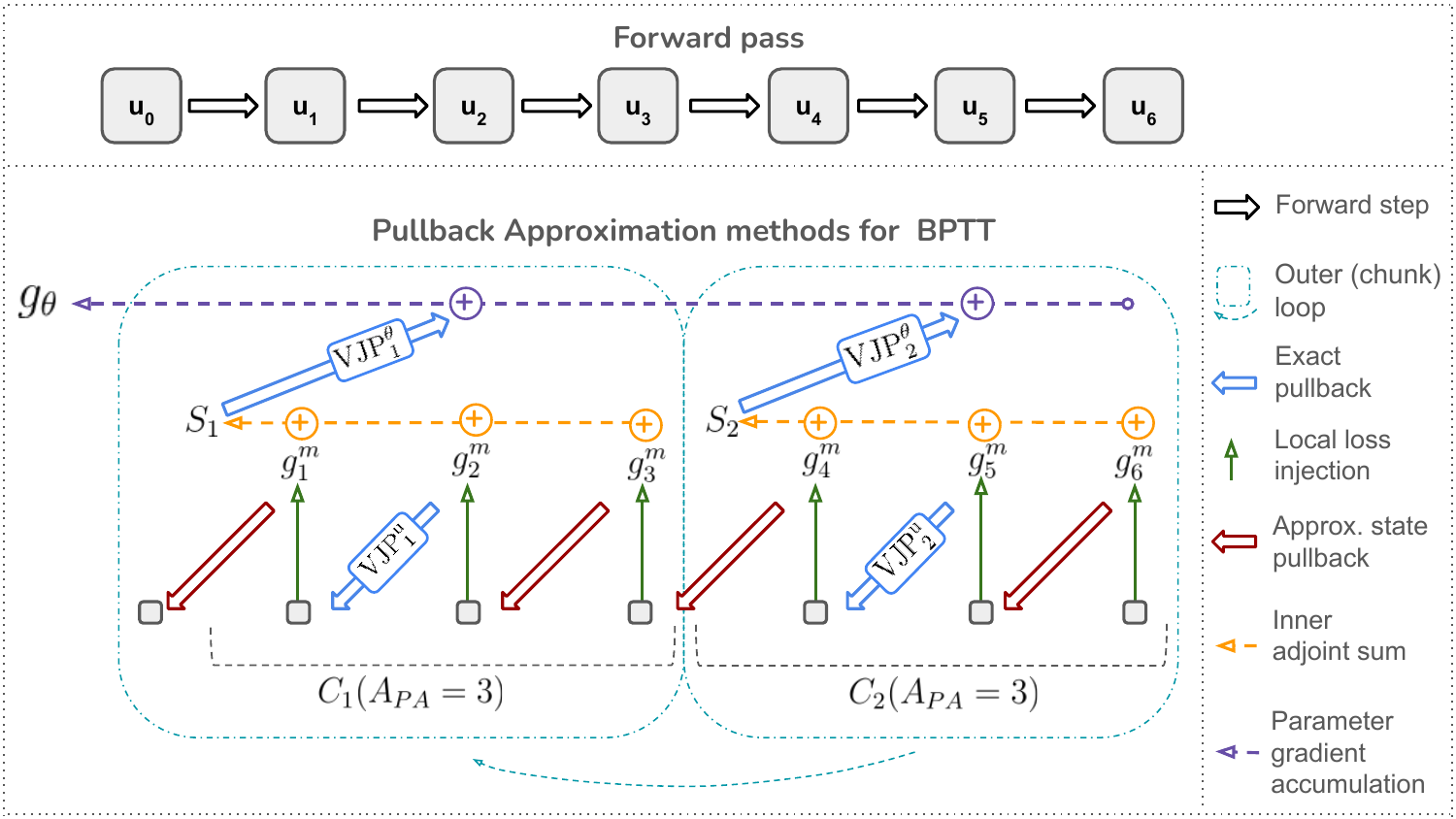}
    \caption{Schematic illustration of the factorized double-loop backward pass
    used by the Zero-order and Pullback Reuse approximations. The trajectory of
    model states $\vect{u}_n$ is partitioned into $K$ chunks of length
    $A_{\text{PA}}$, each containing a single anchor step. During the inner loop,
    the local sensitivity $\vect{g}^{m}_n$ combines the propagated state
    sensitivity with the loss cotangent at step $n$. These step-wise sensitivities
    are accumulated within each chunk into $\vect{S}_i$, which is passed through
    the parameter VJP once per chunk to update the total parameter gradient $\vect{g}_{\vect{\theta}}$.
    }
    \label{fig:grad_approx_schema}
\end{figure}

The corresponding procedure is given in Algorithm~\ref{alg:pullback_approx} and illustrated schematically in Figure~\ref{fig:grad_approx_schema}. Its practical consequence, in terms of computational complexity, is that the number of parameter pullback evaluations no longer scales with the rollout length $N$, but with the number of chunks $K$. Where exact BPTT evaluates one parameter $\text{VJP}^{\theta}$ per time step, the PA methods accumulate the sensitivities into $\vect{S}_i$ through simple vector additions and apply $\text{VJP}^{\theta}$  only once per chunk. The gain is even larger for the ZO method, for which the state pullback is also evaluated only once per chunk, at the anchor step.
\vspace{1em}

\begin{algorithm}[ht]
\small
\caption{Factorized double-loop backward pass for the Zero-order and Pullback Reuse approximations.}
\label{alg:pullback_approx}
\KwIn{anchor states $\{\vect{u}_{n_i^{\text{anc}}}\}_{i=1}^{K}$, terminal sensitivity $\vect{g}_{N+1} = \vect{0}$, loss cotangents $\left\{\frac{\partial \mathcal{L}_N}{\partial \vect{u}_n}\right\}_{n=1}^{N}$}
\KwOut{total parameter gradient $\vect{g}_{\vect{\theta}}$}
$\vect{g} \leftarrow \vect{g}_{N+1}$, \quad $\vect{g}_{\vect{\theta}} \leftarrow \vect{0}$\;
\For{$i = K, \dots, 1$}{
    $(\text{VJP}_i^{\vect{u}}, \text{VJP}_i^{\vect{\theta}}) \leftarrow \text{VJP}\big(f, \vect{u}_{n_i^{\text{anc}}}, \vect{\theta}\big)$ \tcp*{construct the anchor pullbacks}
    $\vect{S}_i \leftarrow \vect{0}$ \tcp*{reset the accumulated chunk sensitivity}
    \For{$n$ in chunk $i$, in reverse order}{
        $\vect{g}^{m}_n \leftarrow \vect{g} + \frac{\partial \mathcal{L}_N}{\partial \vect{u}_n}$ \tcp*{inject the loss cotangent}
        $\vect{S}_i \leftarrow \vect{S}_i + \vect{g}^{m}_n$ \tcp*{accumulate for the parameter VJP}
        \eIf{Zero-order}{
            $\vect{g} \leftarrow \text{VJP}_i^{\vect{u}}(\vect{g}^{m}_n)$ \textbf{if} $n = n_i^{\text{anc}}$, \textbf{else} $\vect{g} \leftarrow \vect{g}^{m}_n$ \tcp*{identity off anchors}
        }{
            $\vect{g} \leftarrow \text{VJP}_i^{\vect{u}}(\vect{g}^{m}_n)$ \tcp*{apply the frozen state VJP}
        }
    }
    $\vect{g}_{\vect{\theta}} \leftarrow \vect{g}_{\vect{\theta}} + \text{VJP}_i^{\vect{\theta}}(\vect{S}_i)$ \tcp*{apply the parameter VJP once per chunk}
}
\Return $\vect{g}_{\vect{\theta}}$\;
\end{algorithm}

\begin{table}[ht]
\centering
\caption{Theoretical complexity of the reverse-mode automatic differentiation
algorithms considered in this work. $N$ denotes the total number of unrolled
time steps, $K$ the number of chunks for the Pullback Approximation methods or
of windows for TBPTT, the latter of length $A_{\mathrm{TB}} = N/K$. A pullback
construction corresponds to one taped forward evaluation of $f$ at the
corresponding linearization state, while a VJP evaluation corresponds to one
application of an already constructed pullback. Residuals denote the
intermediate quantities recorded by automatic differentiation during a single
application of $f$.}
\label{tab:algo_theoretical_complexity}

\footnotesize
\setlength{\tabcolsep}{5pt}
\begin{tabular}{@{}l c c c c@{}}
\toprule
\textbf{Algorithm} &
\begin{tabular}[c]{@{}c@{}}\textbf{Forward} \\ \textbf{model steps}\end{tabular} &
\begin{tabular}[c]{@{}c@{}}\textbf{Pullback} \\ \textbf{constructions}\end{tabular} &
\begin{tabular}[c]{@{}c@{}}\textbf{VJP} \\ \textbf{evaluations}\end{tabular} &
\begin{tabular}[c]{@{}c@{}}\textbf{Peak memory} \\ \textbf{footprint}\end{tabular} \\
\midrule

\textbf{Exact BPTT} & $N$ & $N$ &
$N\,\mathrm{VJP}^{\vect{u}} + N\,\mathrm{VJP}^{\vect{\theta}}$ &
\begin{tabular}[c]{@{}c@{}}$N$ states + \\ $\,N$ AD residuals\end{tabular}
\\[4pt]

\textbf{TBPTT} & $N$ & $N$ &
$N\,\mathrm{VJP}^{\vect{u}} + N\,\mathrm{VJP}^{\vect{\theta}}$ &
\begin{tabular}[c]{@{}c@{}}$A_{\mathrm{TB}}$ states + \\ $\,A_{\mathrm{TB}}$ AD residuals\end{tabular}
\\[4pt]

\textbf{Zero-Order} & $N + K$ & $K$ &
$K\,\mathrm{VJP}^{\vect{u}} + K\,\mathrm{VJP}^{\vect{\theta}}$ &
\begin{tabular}[c]{@{}c@{}}$N$ states + \\ $\,1$ AD residual\end{tabular}
\\[4pt]

\textbf{Pullback Reuse} & $N + K$ & $K$ &
$N\,\mathrm{VJP}^{\vect{u}} + K\,\mathrm{VJP}^{\vect{\theta}}$ &
\begin{tabular}[c]{@{}c@{}}$N$ states +\\ $\,1$ AD residual\end{tabular}
\\[4pt]

\textbf{$\sqrt{N}$ Checkpointing} & $2N$ & $N$ &
$N\,\mathrm{VJP}^{\vect{u}} + N\,\mathrm{VJP}^{\vect{\theta}}$ &
\begin{tabular}[c]{@{}c@{}}$\sqrt{N}$ states +\\ $\,\sqrt{N}$ AD residuals\end{tabular}
\\

\bottomrule
\end{tabular}
\end{table}
The theoretical computational cost and memory requirements of the proposed algorithms are reported in Table~\ref{tab:algo_theoretical_complexity}, alongside those of standard exact BPTT, TBPTT and $\sqrt{N}$ checkpointing \citep{chen_etal_2016_sqrt_checkpointing}. The table summarizes the requirements of each algorithm, independently of the hardware on which it is run. The number of forward model steps includes both the initial forward pass together with the recomputations required to contruct the anchor pullbacks. The VJP counts  distinguish the state and parameter vector--Jacobian products $\text{VJP}^{\vect{u}}$ and $\text{VJP}^{\vect{\theta}}$.The peak memory estimates report the number of model states held simultaneously, together with the number of active AD residuals, that is, the intermediate quantities recorded by automatic differentiation during a single application of the time-stepping map $f$. Overall, Table~\ref{tab:algo_theoretical_complexity} shows that the PA methods substantially reduce memory requirements, by retaining a single residual at any time, while also reducing the number of pullback constructions and VJP evaluations during the backward pass.

\section{Related works}\label{sec:related_works}

\textbf{Forward-mode versus reverse-mode differentiation.} Parameter sensitivities can be propagated in either direction. Forward mode carries one tangent per parameter direction alongside the state, so nothing is stored and memory stays constant in the rollout length, but its cost grows linearly with the number of parameters, which rules it out for the neural sub-models considered here. Reverse mode runs the trajectory first and propagates the adjoint back through it: the cost becomes largely independent of the parameter count, at the price of retaining the computational graph \citep{sapienza_etal_2025_diff_progra}. For recurrent models, TBPTT \citep{williams_and_peng_1990_tbptt, william_and_zipser_1995_tbptt} and its unbiased variant \citep{tallec_&_yann_2017_unbiased_bptt} address the latter by truncating that graph. We instead keep the reverse-mode formulation intact and approximate the operators within it.

\textbf{Gradient checkpointing.} The standard way of reducing the memory footprint of reverse-mode differentiation is checkpointing, in which only a subset of the forward states is retained and the missing segments are recomputed on the fly during the backward pass. The $\sqrt{N}$ strategy of \citet{chen_etal_2016_sqrt_checkpointing} stores $\sqrt{N}$ checkpoints, bringing memory down to $\mathcal{O}(\sqrt{N})$ for the price of one extra forward pass; optimal schedules for a prescribed memory budget were derived earlier in the numerical-analysis literature \citep{griewank_and_walther_2000_checkpointing_revolve}, and the idea has since been carried over to recurrent architectures \citep{gruslys_etal_2016_memory_efficient_rnn_backprop}. Checkpointing trades memory for computation while leaving the gradient exact: every pullback is still constructed and every VJP still evaluated. Conversely, the methods proposed here reduce both quantities instead, at the price of a gradient that is no longer exact. 

\textbf{Gradient approximation for non-differentiable operations.} Replacing a Jacobian by the identity is a long-standing practice in deep learning. The straight-through estimator \citep{bengio_etal_2013_grad_conditional_computation} propagates the incoming gradient unchanged through quantization or other non-differentiable operations, and similar surrogate backward passes are used for discrete sampling \citep{jang_etal_2017_gumbel_softmax_reparam} and binarized networks \citep{hubara_etal_2016_binarized_nn}. These estimators are biased by construction, yet they remain effective for optimization, which is precisely the behavior we observe for the ZO method. The setting however differs: in deep learning, the identity substitutes a Jacobian that does not exist, whereas here it approximates a Jacobian that exists and is close to the identity for small time steps, so the approximation carries an order of accuracy in $h$ rather than purely heuristic justification.

\textbf{Jacobian reuse} Freezing a Jacobian over several iterations is standard practice in numerical analysis. Modified Newton and quasi-Newton methods reuse a factorization computed at an earlier iterate rather than re-evaluating it at every step \citep{broyden_1965_methods_nonlinear_eqs, alexander_1991_modified_newton_stiff_odes}, and implicit time-integration solvers refresh the Jacobian only periodically, balancing the cost of its evaluation against the convergence rate of the inner solve \citep{gustafsson_and_soderlind_control_iter_sol_nonlinear_eqs_ode_solvers, hindmarsh_etal_2005_sundials}. Pullback Reuse transposes this practice to the backward pass of an unrolled trajectory, reusing the anchor state pullback across a chunk instead of reconstructing it at every step. The consequence of a stale linearization is however not the same in the two settings: an implicit solve is self-correcting, in that a Jacobian evaluated away from the current iterate slows convergence but leaves the converged solution exact, whereas a backward pass has no inner iteration to absorb the error, so a frozen pullback propagates into the parameter gradient as a bias. This is what motivates the gradient-fidelity diagnostics reported in Section \ref{sec:experimental_setup_results}.

\section{Experimental setup, analyses and results}\label{sec:experimental_setup_results}

We evaluate the gradient approximation methods on three systems of increasing complexity: the Lorenz-63 system, a two-dimensional large-eddy simulation of quasi-geostrophic turbulence, and a semi-realistic multilayer quasi-geostrophic model of the western Mediterranean Sea. Each case study is presented in turn, with its configuration followed directly by its results.

For each case study we conduct two complementary analyses. The first quantifies how accurately each approximation reproduces the exact gradient, and at what computational cost. Gradient fidelity is measured by the cosine similarity with the exact gradient: since gradient-based optimization is driven by the descent direction rather than by the gradient magnitude, this quantifies how well an approximation preserves the optimization direction. All methods are implemented in JAX; peak memory is estimated from XLA’s post-compilation buffer analysis, and execution time from synchronized runs of the compiled functions.

The second assesses whether the approximate gradients remain sufficiently informative to train hybrid models. The hybrid models are trained by minimizing an online objective function of the form \eqref{eq:optimization_global}, based on the mean squared error of the simulated state variables over a rollout of length N. The exact formulation of the online loss for each case study is given in \ref{app:online_objectives}.

Both analyses are carried out for several degrees of approximation $\alpha$. For TBPTT, $\alpha= A_{TB}$ is the window length; for the Pullback Approximation methods, $\alpha= K$ is the number of chunks, and hence of anchor points. In both cases a smaller $\alpha$ means a more aggressive approximation.  The two families are aligned by degree of approximation rather than by partition size.

\subsection{Lorenz-63}

The Lorenz-63 (L63) system is a simplified model of atmospheric convection consisting of three coupled, first-order nonlinear differential equations \eqref{eq:classic_l63}, where the state vector is given by $\vect{u}_t = [x_t, y_t, z_t]$ and $\sigma$, $\rho$ and $\beta$ are positive constants. We use the standard values $\sigma = 10$, $\rho = 28$ and $\beta = 8/3$, for which the system is chaotic and its trajectories converge to the Lorenz attractor \citep{lorenz_1963_l63}.

The hybrid setup follows \eqref{eq:ode_system}. The dynamical model $\vect{F}$ is an incomplete L63 system \eqref{eq:incomplete_l63} in which the term $-\beta z_t$ has been removed which prevents the model from reproducing the true attractor. A sub-model $\mathbf{M}(\mathbf{u}_t,\vect{\theta}) = [m1(\cdot),m2(\cdot),m3(\cdot)]^T$ is then coupled to it \eqref{eq:corrected_l63} to compensate for the missing dynamics. Since the source of model error is unknown in realistic settings, the correction is applied to all three prognostic equations rather than to the modified one alone.
\vspace{1em}

\begin{subequations}
\label{eq:l63_all}
\begin{minipage}{0.25\textwidth}
    \centering
    \textbf{Classic L63 \newline(Ground truth)}
    \begin{equation}
    \label{eq:classic_l63}
    \begin{aligned}
        \dot{x}_t &= \sigma (y_t - x_t), \\
        \dot{y}_t &= x_t (\rho - z_t) - y_t, \\
        \dot{z}_t &= x_t y_t - \beta z_t
    \end{aligned}
    \end{equation}
\end{minipage}
\hfill
\begin{minipage}{0.25\textwidth}
    \centering
    \textbf{Incomplete L63}
    \begin{equation}
    \label{eq:incomplete_l63}
    \begin{aligned}
        \dot{x}_t &= \sigma (y_t - x_t), \\
        \dot{y}_t &= x_t (\rho - z_t) - y_t, \\
        \dot{z}_t &= x_t y_t
    \end{aligned}
    \end{equation}
\end{minipage}
\hfill
\begin{minipage}{0.32\textwidth}
    \centering
    \textbf{Hybrid L63}
    \begin{equation}
    \label{eq:corrected_l63}
    \begin{aligned}
        \dot{x}_t &= \sigma (y_t - x_t) + {m}_1(\vect{u}_t, \theta), \\
        \dot{y}_t &= x_t (\rho - z_t) - y_t + {m}_2(\vect{u}_t, \theta), \\
        \dot{z}_t &= x_t y_t - \beta z_t + {m}_3(\vect{u}_t, \theta)
    \end{aligned}
    \end{equation}
\end{minipage}
\end{subequations}
\vspace{1em}

The sub-model is a fully connected multilayer perceptron with a single hidden layer of 32 units and ReLU activations, mapping the three-dimensional state to a three-dimensional correction added to the model tendency. The reference data are generated with the true system \eqref{eq:classic_l63} after a spin-up of 5 L63 time units, so that the trajectory lies on the attractor. We sample $n = 2{,}000$ points at $dt = 0.01$, corresponding to $T = 20$ time units, and use the first 70\% for training and the last 20\% for testing; the intermediate 10\% is discarded to avoid temporal leakage. The training setup is summarized in the \ref{app:model_params} Table~\ref{tab:training_setup_l63}.

\subsubsection{Gradient fidelity and computational cost}\label{subsec:grad_fid_lorenz}

\begin{figure}[ht]
    \centering
    \includegraphics[width=1\linewidth]{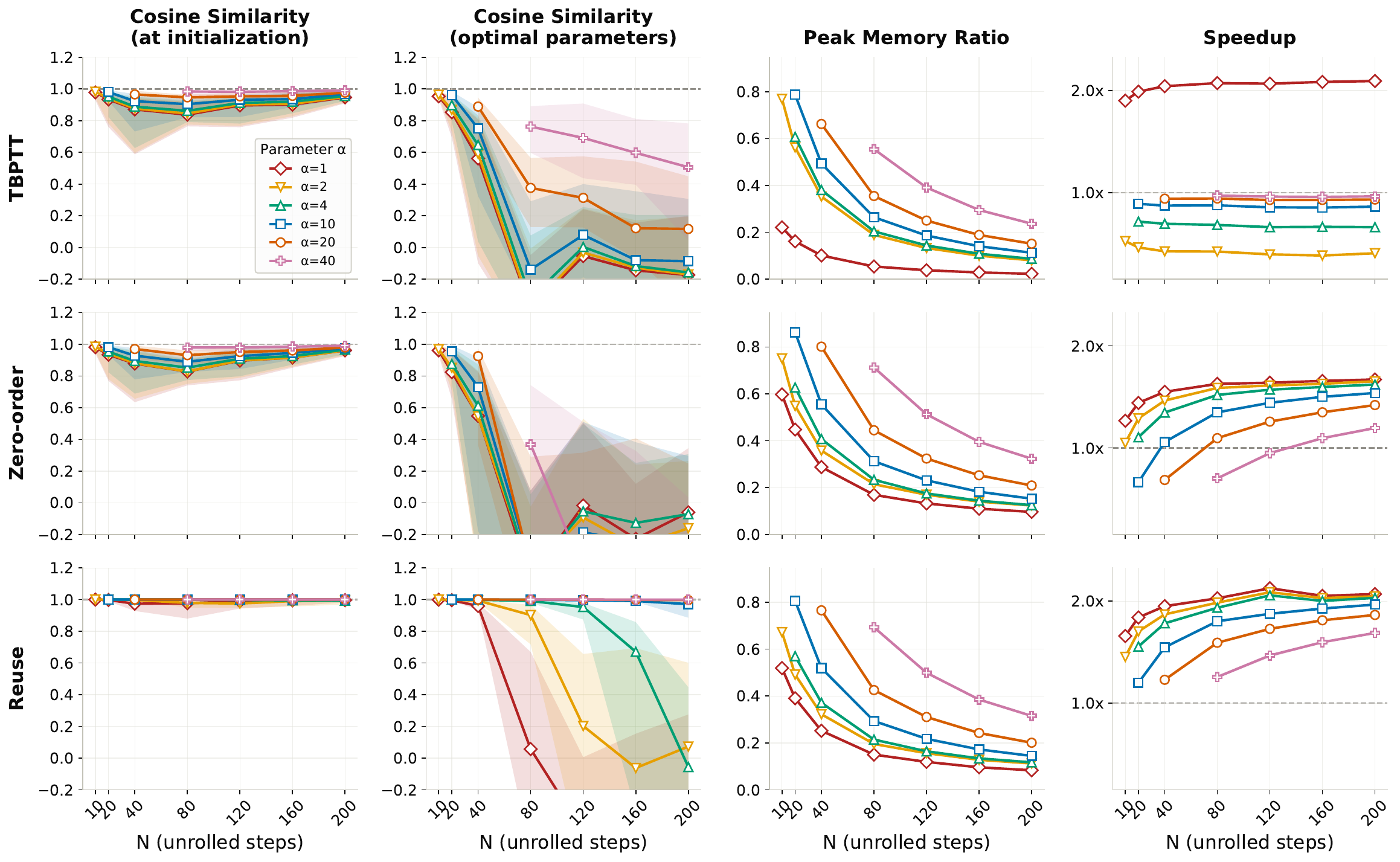}
        \caption{Gradient fidelity and computational performance comparison of the approximate gradient methods relative to exact BPTT for the hybrid L63 model (incomplete L63 plus sub-model correction. Cosine similarity is computed with respect to the exact gradient, both at initialization (pre-training) and after successful training. Parameter $\alpha$ controls the degree of approximation, with smaller values corresponding to a more aggressive approximation.}
\label{fig:L63_grad_benchmark}
\end{figure}

Figure~\ref{fig:L63_grad_benchmark} summarizes the fidelity and cost of TBPTT, ZO and Reuse relative to exact BPTT, across increasing unrolled trajectory lengths $N$ and degrees of approximation $\alpha$. The results reveal a trade-off between the two, with gradient fidelity depending on both quantities.

Before training, all three methods track the exact gradient direction closely, with cosine similarity above 0.8 over the full range tested, and Reuse remains almost perfectly aligned throughout. After training, the agreement breaks down unevenly and TBPTT and ZO degrade rapidly as $N$ increases and $\alpha$ decreases, while Reuse stays close to 1 over a much wider range. 

This contrast can be understood dynamically. Before training, the incomplete model collapses onto a fixed point rather than the true L63 attractor, so the local Jacobians vary little from step to step and even a coarse approximation remains accurate. After training, the hybrid model recovers the full attractor and the trajectory becomes chaotic, so the adjoint is subject to exponential divergence that scales with the rollout length $N$. In that regime, retaining some curvature information within each chunk, as Reuse does through its frozen anchor Jacobian, preserves the descent direction far more reliably than truncating the recurrence (TBPTT) or discarding the local state sensitivity transformation (ZO).

Regarding peak memory and runtime, the memory ratio falls with N, since the footprint of exact BPTT grows faster, and rises with $\alpha$, as larger windows (TBPTT) and more anchors (PA) both retain more information in the tape. Runtime separates the methods more clearly. TBPTT repeats the same forward and backward passes as exact BPTT, so no speedup is expected, and none is observed for $\alpha > 1$; the apparent $2\times$ speedup at $\alpha = 1$ is an implementation effect,which we attribute to the compiler fusing backward-pass operations, and thereby avoiding the storing and reloading of intermediate states. Such an effect is only visible when a single step time step costs little against the loop overhead, as for L63, but not for the quasi-geostrophic system of Section~\ref{subsec:QG}. ZO and Reuse both improve with $N$ and are fastest at the fewest anchors. Reuse is consistently ahead of ZO here, despite the latter
saving additional $\text{VJP}^{\vect{u}}$ evaluations. With three state variables and a compact MLP, a VJP evaluation is cheap, so the conditional branch ZO applies at anchor steps costs more than the operations it avoids, which is no longer true for the larger quasi-geostrophic system in Section~\ref{subsec:QG}.

\subsubsection{Training with approximate gradients}\label{subsec:training_lorenz}

The analyses above characterize the intrinsic properties of the approximations in terms of gradient accuracy and computational efficiency. In this section, we evaluate whether the resulting gradients remain informative enough to optimize a hybrid model. In this context, we train identical models with exact BPTT, TBPTT, ZO and Reuse under identical optimization settings.

\begin{figure}[H]
    \centering
    \includegraphics[width=1\linewidth]{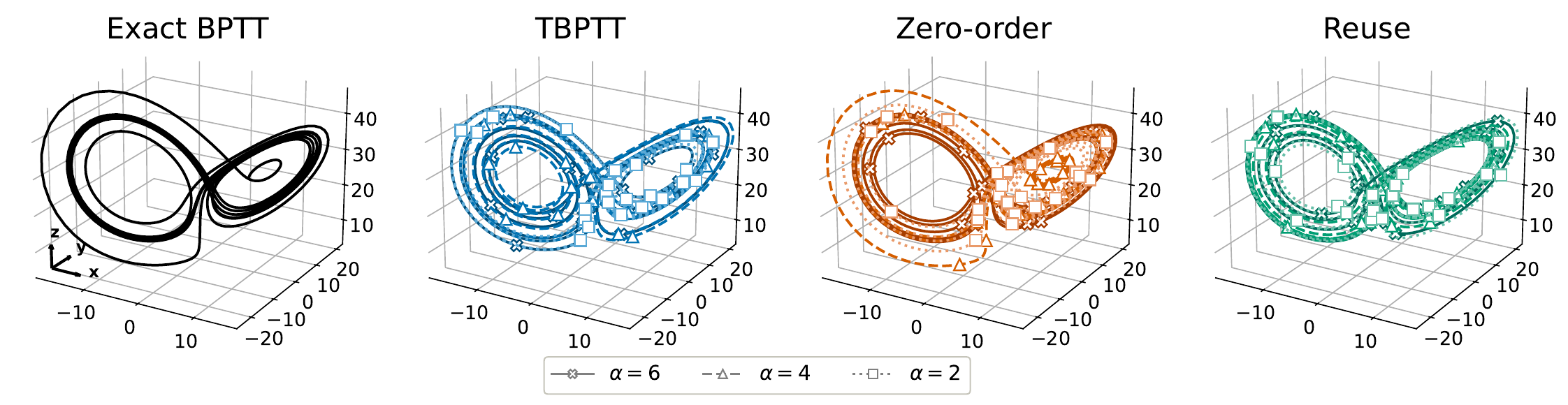}
    \caption{Lorenz-63 trajectories generated by the post-training models over 2 Lorenz-63 time units (2,000 time steps). The models were trained for 250 epochs with a training window of $N=12$ unrolled time steps. Results are shown for Exact BPTT, TBPTT, Zero-Order, and Reuse Gradient, with the latter three evaluated at $\alpha=2$, $4$, and $6$.}
    \label{fig:trained_l63_diff_grad_approx1}
\end{figure}

Comparable losses do not necessarily imply comparable learned dynamics, particularly for a chaotic system, so we also evaluate the trained models in phase space. Figure~\ref{fig:trained_l63_diff_grad_approx1} shows that all methods, down to the the configurations with the fewest anchors, recover the characteristic two-lobed geometry of the Lorenz attractor, even though individual trajectories diverge. The approximate gradients are therefore sufficient for the optimizer to recover a correction that reproduces the right dynamics, despite following different optimization paths.
\begin{figure}[H]
    \centering
    \includegraphics[width=1\linewidth]{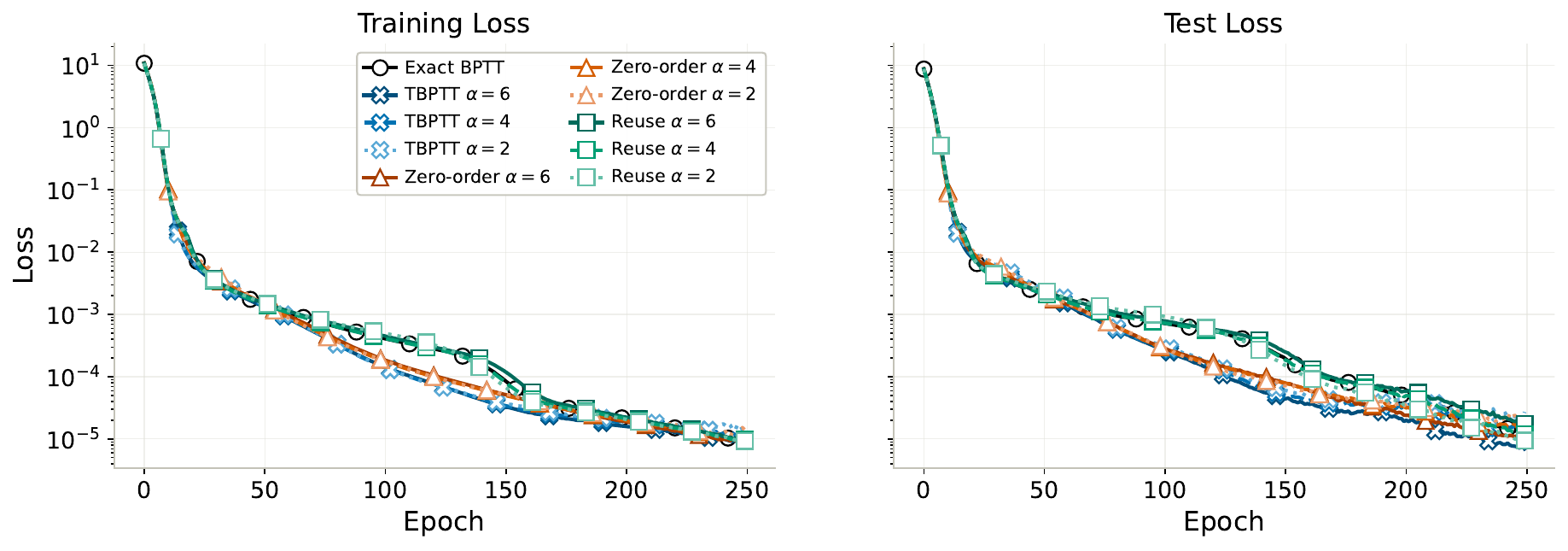}
     \caption{Evolution of the training and test losses over 250 epochs for the exact BPTT, TBPTT, Zero-order, and Pullback Reuse gradient methods with $\alpha=6$, $\alpha=4$ and $\alpha=2$. The MLP is trained using a training window of N=12 unrolled timesteps.}
    \label{fig:gradient_methods_loss_curves_l63}
\end{figure}

\subsection{Large eddy simulation of quasi-geostrophic turbulence}\label{subsec:QG}

In this second experiment, we consider Quasi-Geostrophic (QG) dynamics, a workhorse of geophysical fluid dynamics, valid where the flow is hydrostatic and the Coriolis acceleration balances the horizontal pressure gradient. 

Over a doubly periodic $(x, y)$ square domain with a side length $L=2 \pi$, the dimensionless governing equations for vorticity $(\omega_t)$ and streamfunction $(\psi_t)$ are
\begin{subequations}
\label{eq:QG_equations_true}
\begin{align}
\frac{\partial \omega_t}{\partial t}+\mathcal{A}(\omega_t, \psi_t) & =\frac{1}{\operatorname{Re}} \nabla^2 \omega_t-f-r \omega_t \label{eq:equation_vorticity} \\
\nabla^2 \psi_t & =-\omega_t \label{eq:equation_stream_vorticity}
\end{align}
\end{subequations}
where, $\mathcal{A}(\omega_t, \psi_t)$ is the nonlinear advection term:
$$
\mathcal{A}(\omega_t, \psi_t)=\frac{\partial \psi_t}{\partial y} \frac{\partial \omega_t}{\partial x}-\frac{\partial \psi_t}{\partial x} \frac{\partial \omega_t}{\partial y},
$$
and $f$ is a deterministic forcing
$$
f(x, y)=k_f\left[\cos \left(k_f x\right)+\cos \left(k_f y\right)\right] \text {. }
$$

We consider a large eddy simulation (LES) setting in which both fields are filtered with a Gaussian filter \citep{sagaut_and_levasseur_2005_multiscale_LES_turbulence}, denoted $\overline{(\cdot)}$. Applying the filter to \eqref{eq:equation_vorticity}--\eqref{eq:equation_stream_vorticity} yields
\begin{subequations}
\label{eq:QG_LES}
\begin{align}
\frac{\partial \bar{\omega}_t}{\partial t}+\mathcal{A}(\bar{\omega}_t, \bar{\psi}_t) & =\frac{1}{\operatorname{Re}} \nabla^2 \bar{\omega}_t-\bar{f}-r \bar{\omega}_t+\underbrace{\mathcal{A}(\bar{\omega}_t, \bar{\psi}_t)-\overline{\mathcal{A}(\omega_t, \psi_t)}}_{\Pi_t \approx \vect{M}_{\vect{\theta}}} \label{eq:equation_vorticity_LES} \\
\nabla^2 \bar{\psi}_t & =-\bar{\omega}_t \label{eq:equation_stream_vorticity_LES}
\end{align}
\end{subequations}
Solving the LES at a coarser resolution than a direct numerical simulation (DNS) of \eqref{eq:equation_vorticity}-\eqref{eq:equation_stream_vorticity} is considerably cheaper, but requires a closure for the subgrid-scale (SGS) term $\Pi_t$. Here $\Pi_t$ is parameterized by a convolutional neural network (CNN)  $\vect{M}_{\vect{\theta}}(\bar{\psi}_t, \bar{\omega}_t, \vect{\theta})$ that takes as inputs both the vorticity and streamfunction fields $(\bar{\psi}_t, \bar{\omega}_t)$. Together, the LES solver and the CNN sub-model form  a hybrid model of the form introduced in Section \ref{sec:problem_statement}

The reference DNS trajectory is initialized using a random Gaussian vorticity field on a 1024 x 1024 grid, and spun up for 200,000 steps to reach statistical equilibrium, after which a trajectory of 2,000,000 steps is generated with a timestep $h_{DNS} = 5 \times 10^{-5}$. The target dataset is obtained by spatially  filtering this trajectory with Gaussian filter of  width $\Delta_f = 2\Delta_{LES}$, coarse-graining to a 64 x 64 LES grid, and sampling at the LES timestep $h_{LES} = 8 \times 10^{-4}$, that is one snapshot every 16 DNS steps. The resulting dataset is split into 50 physical time units for training and 20 for validation.

The CNN is trained over a rollout of $N = 40$ steps. In what follows we benchmark the gradient approximation methods against the exact BPTT gradient baseline, to assess their efficacy and stability on this online learning problem.

\subsubsection{Gradient fidelity and computational cost}\label{subsec:grad_fid_qg}

\begin{figure}[H]
    \centering
    \includegraphics[width=1\linewidth]{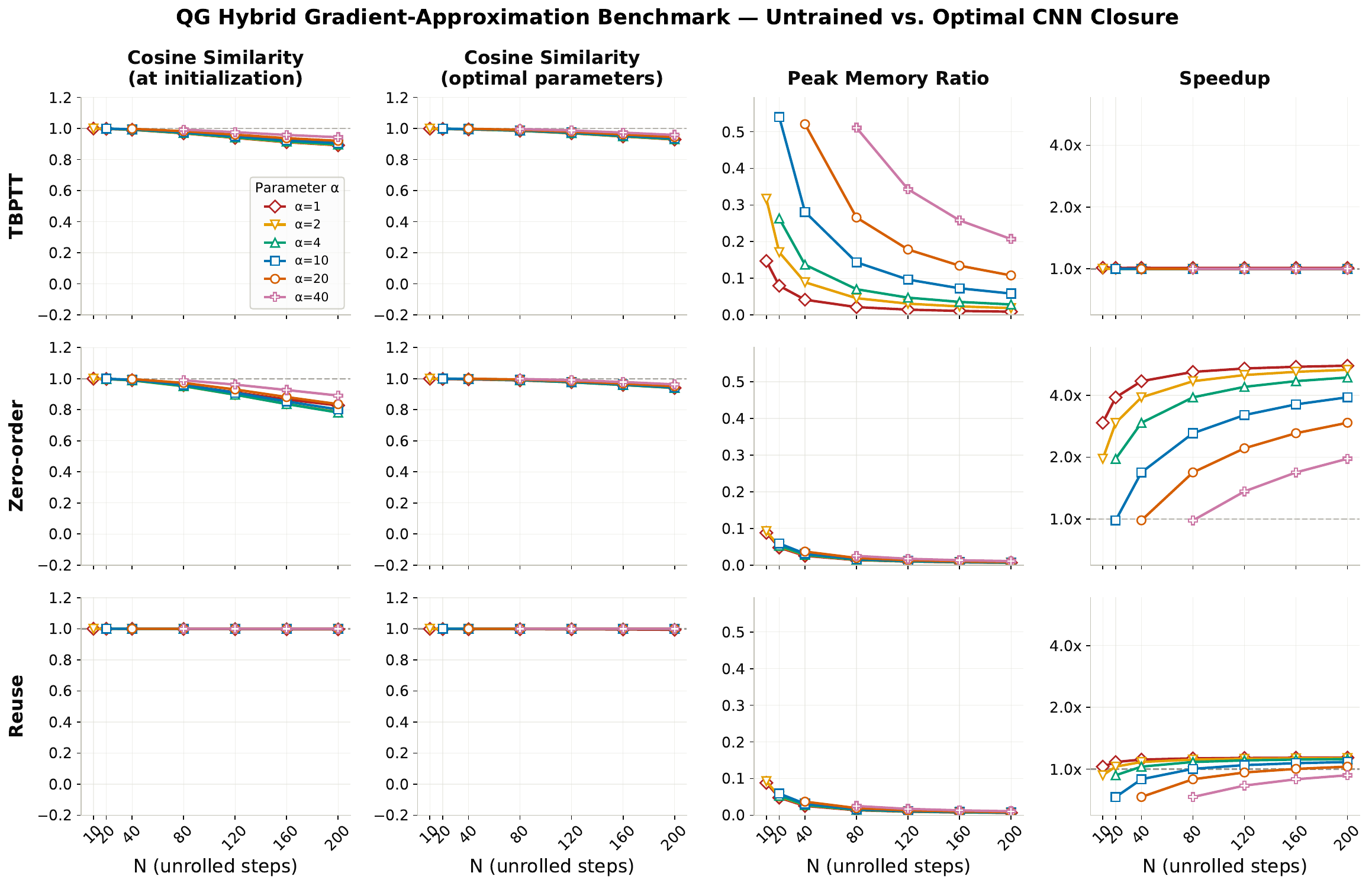}
    \caption{Gradient fidelity and computational performance of the approximate gradient methods relative to Exact BPTT, for the double-periodic single-layer hybrid QG model (LES plus CNN closure). Cosine similarity is computed with respect to the exact gradient, both at initialization (pre-training) and after successful training. For TBPTT, $\alpha$ denotes the chunk size; for Zero-Order and Reuse, it denotes the number of chunks (anchor points). In all cases, $\alpha$ controls the degree of approximation.}
    \label{fig:2DQG_grad_benchmark}
\end{figure}

The cosine similarity reported in Figure~\ref{fig:2DQG_grad_benchmark} remains high for all tested methods.
TBPTT and ZO degrade gradually as the unrolling length $N$ increases, whereas Reuse stays almost perfectly aligned with the exact gradient at every degree of approximation $\alpha$. Suggesting that for QG dynamics, the level of approximation could therefore be pushed further without noticeably degrading the gradient.

The separation in peak memory between TBPTT and the PA methods is much clearer here than for L63. Each model state and its associated residuals are far larger in this higher-dimensional system, since a single application of $f$ involves spectral transforms and the intermediate Runge-Kutta stages, so differences in the amount of active backward-pass information show up directly. ZO and Reuse retain a single active residual at a time, whereas TBPTT keep the states and residuals of its entire truncation window, and the measured memory ratio grows with the truncation window with $\alpha$ accordingly. Notably, TBPTT at $\alpha = 1$ approximately coincides with the PA methods, confirming that the residuals account for essentially all of the memory these methods save.

Runtime separates the methods further. TBPTT provides essentially no speedup over exact BPTT.  Truncation reduces what is held in memory at once, but the same sequence of state and parameter pullbacks is still evaluated over the full trajectory, so the memory saving does not translate into time. Reuse is slightly faster  at the most aggressive approximation levels ($\alpha \leq 10$) and a slower elsewhere, reflecting two competing effects; the saving from adjoint factorization of Section~\ref{subsec:num_algo}, and the cost of constructing a pullback at each anchor. Few anchors means few constructions and the factorization dominates;  with many anchors the recomputation overhead outweighs it. ZO instead skips the state VJP entirely between anchors, reaching a speedup of $\sim$ 4 to 5$\times$ over exact BPTT at the most aggressive $\alpha$. Read alongside the L63 results in Section \ref{subsec:grad_fid_lorenz}, this confirms that the extra $VJP^{\vect{u}}$ evaluations retained by Reuse dominate once the state dimension is large enough for a VJP to be expensive.

\subsubsection{Training with approximate gradients}\label{subsec:training_qg}

\begin{figure}[ht]
    \centering
    \includegraphics[width=1\linewidth]{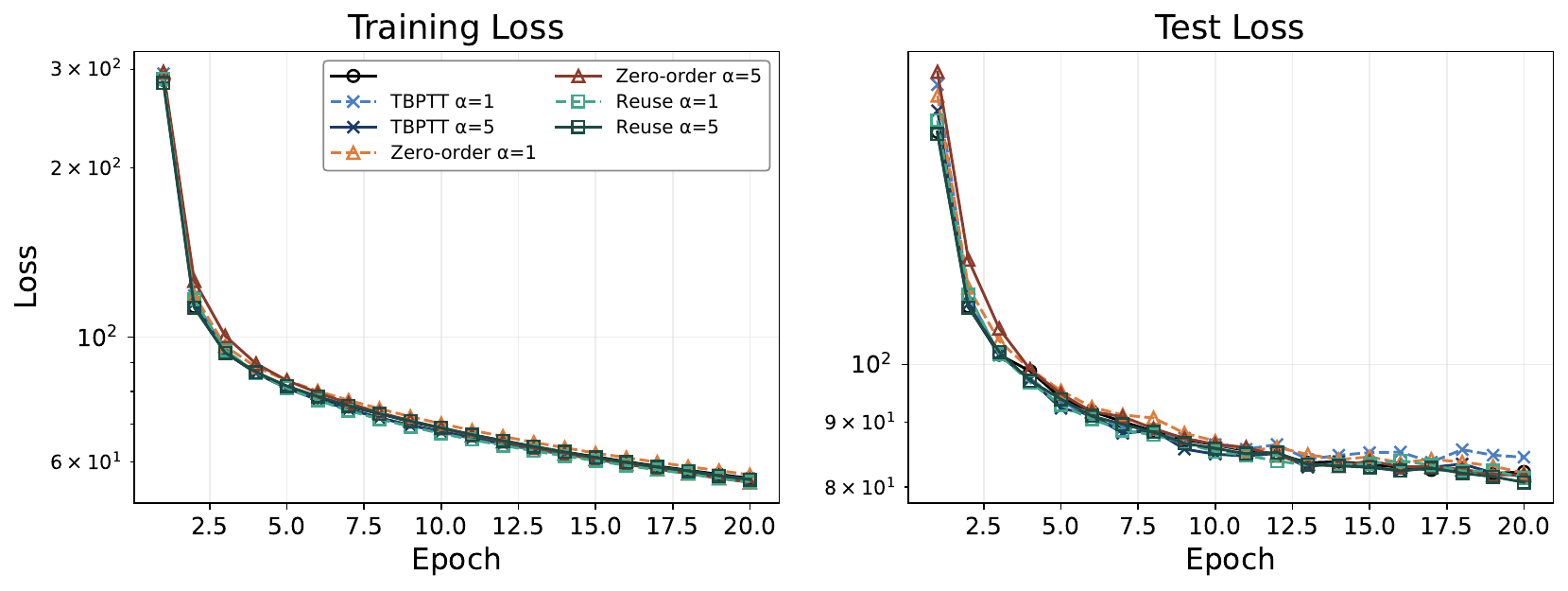}
    \caption{Evolution of the training and test losses over 20 epochs for the Exact BPTT, TBPTT, Zero-Order, and Reuse Gradient methods with $\alpha=1$ and $\alpha=5$. The CNN is trained using a window of N=40 unrolled timesteps.}
    \label{fig:train_val_loss_2DQG}
\end{figure}

\begin{figure}[H]
    \centering
    \includegraphics[width=0.80\linewidth]{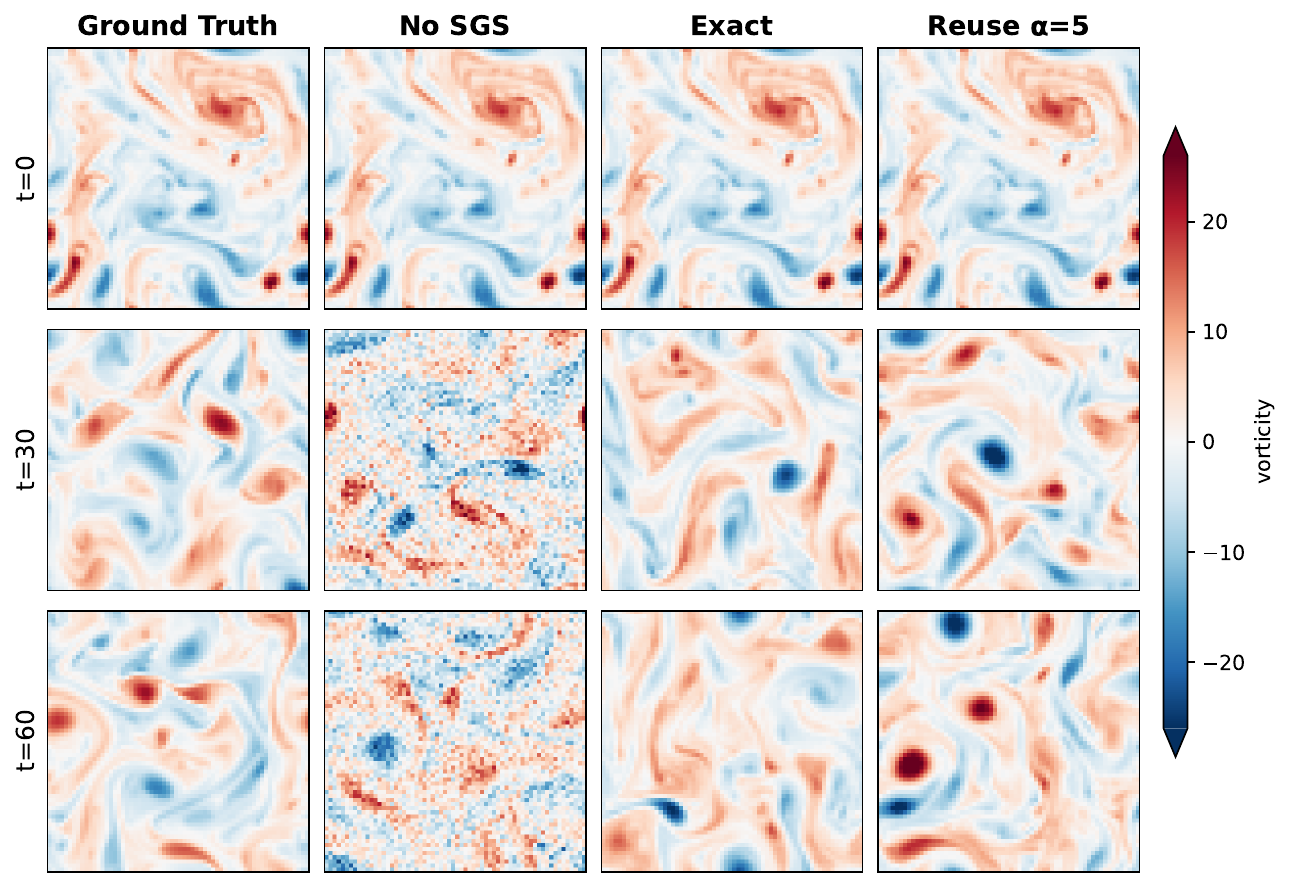}
    \caption{Post-training evolution of the vorticity field at selected times ($t=0$--$60$) for the Ground Truth, No SGS, Exact, and Reuse ($
    \alpha=5$) simulations. The CNN was trained for 20 epochs using a window of $N=40$ unrolled time steps.}
    \label{fig:2DQG_snap}
\end{figure}

Figure~\ref{fig:train_val_loss_2DQG} shows the training and test losses. All methods follow the same convergence pattern, both losses decreasing rapidly toward a common low-loss regime. The optimization paths separate slightly during parts of training, but these differences are small relative to the overall reduction in loss, and convergence does not degrade systematically as $\alpha$ decreases. This is consistent with the cosine similarities of Figure~\ref{fig:2DQG_grad_benchmark}, where every method stays closely aligned with the exact gradient at all approximation levels.
We further evaluate the trained models in free-running long-term simulations. Figure~\ref{fig:2DQG_snap} shows the post-training evolution of the vorticity field for the models trained with the exact BPTT and the Reuse gradients, together with the No SGS simulation;  the other methods behave similarly and are not shown. Without a suggrid-scale closure, the simulation progressively develops spurious small-scale, spatially and incoherent vorticity fluctuations. Both hybrid models instead maintain coherent structures whose spatial scales and vorticity magnitudes match the reference throughout the simulated interval. The two decorrelate progressively, as expected for a chaotic system, but their overall organization and order of magnitude remain comparable to the ground truth. The Reuse gradient is therefore sufficient to maintain a stable correction and to recover the resolved-scale dynamics from sparser gradient information than exact BPTT.

\subsection{Bias correction of a multilayer quasi-geostrophic ocean model}\label{subsec:methods_ocean}

To evaluate the proposed methods in a more realistic setting, we take the dynamical core $\vect{F}$ to be a multilayer quasi-geostrophic (GC) ocean model. We use a JAX implementation of MQGeometry \citep{thiry_mqgeometry-10_2024}, a differentiable multilayer QG solver based on the formulation of \cite{thiry_unified_2024}. Rather than evolving the potential vorticity, MQGeometry recasts the QG equations as a projection of the multilayer rotating shallow water (RSW) equations, with the state expressed in terms of the horizontal velocities and the layer-thickness anomaly, $\vect{u}_t = [u_t, v_t, h_t]^{\top}$. The system evolves under the full nonlinear RSW operator $\vect{\varphi}$ while the projection $P$ onto the geostrophic manifold filters fast ageostrophic motions, $\dot{\vect{u}}_t = P \circ \vect{\varphi}(\vect{u}_t)$. Advection uses high-order upwind-biased WENO reconstructions, whose implicit dissipation removes the need for explicit hyperviscosity. Full details are given in \citet{thiry_unified_2024, thiry_mqgeometry-10_2024}.

The configuration is a regional western Mediterranean domain spanning 31°N--45°N and 5.5°W--9°E, discretized on a $310 \times 254$ grid at approximately 5~km resolution. The water column is represented by three layers of 75~m, 400~m and 2525~m respectively, denoted Surface, Middle and Bottom, and chosen to match the typical stratification of the area \citep{millot_1999_west_med_sea}. Bathymetry is only used to define the continental boundaries and to exclude regions shallower than 100~m. Surface forcing is prescribed from a monthly mean winds for 2010, converted to wind stress with a constant drag coefficient and interpolated in space and time; bottom stress follows a linear friction law with constant coefficient $C_D$. Physical and numerical parameters are summarized in \ref{app:model_params}, Table~\ref{tab:ocean_reg_config}.

The structural bias is introduced by amplifying the bottom drag coefficient $C_D$ by a factor of five. Bottom drag enters the bottom-layer momentum tendencies as a linear damping of the horizontal velocities, and its effect propagates upward through coupling between baroclinic modes, so that relatively small changes in $C_D$ produce significant differences in kinetic energy and circulation. We restore the correct dynamics in two ways, by calibrating $C_D$ directly and by correcting the QG tendencies with a neural network.

In the first strategy, $C_D$ is treated as a learnable scalar and calibrated by gradient descent on $\frac{d\mathcal{L}_N}{dC_D}$. No sub-model is involved, so this is not a hybrid setup, but it remains an online problem and provides a transparent baseline for the convergence and fidelity of the proposed methods. In the second strategy, a sub-model $\mathbf{M}(\vect{u}_t, \vect{\theta})$ is coupled to the biased model following \eqref{eq:ode_system} and learns to correct its tendencies. Since the source of model error is unknown in realistic settings, the correction is applied to all prognostic equations and all three vertical layers rather than to the bottom-layer momentum alone. The sub-model is a compact CNN with two convolutional layers of 64 and 32 feature maps, both with $3 \times 3$ kernels followed by ReLU activations, and a final $1 \times 1$ convolution mapping to the output channels.

The true dynamics are spun up from rest over twenty years, with the domain-averaged kinetic energy monitored until equilibrium. The final spin-up state initializes a five-year ground-truth simulation, saved every two hours, of which 70\% is used for training and 20\% for testing, the intermediate 10\% being discarded to avoid temporal leakage. Training uses a rollout of $N$ = 12 steps. The remaining hyperparameters are given in \ref{app:model_params}, Table~\ref{tab:training_setup}.

\subsubsection{Parameter calibration}\label{subsubsec:results_ocean_cd}

\begin{figure}[H]
    \centering
\includegraphics[width=1\linewidth]{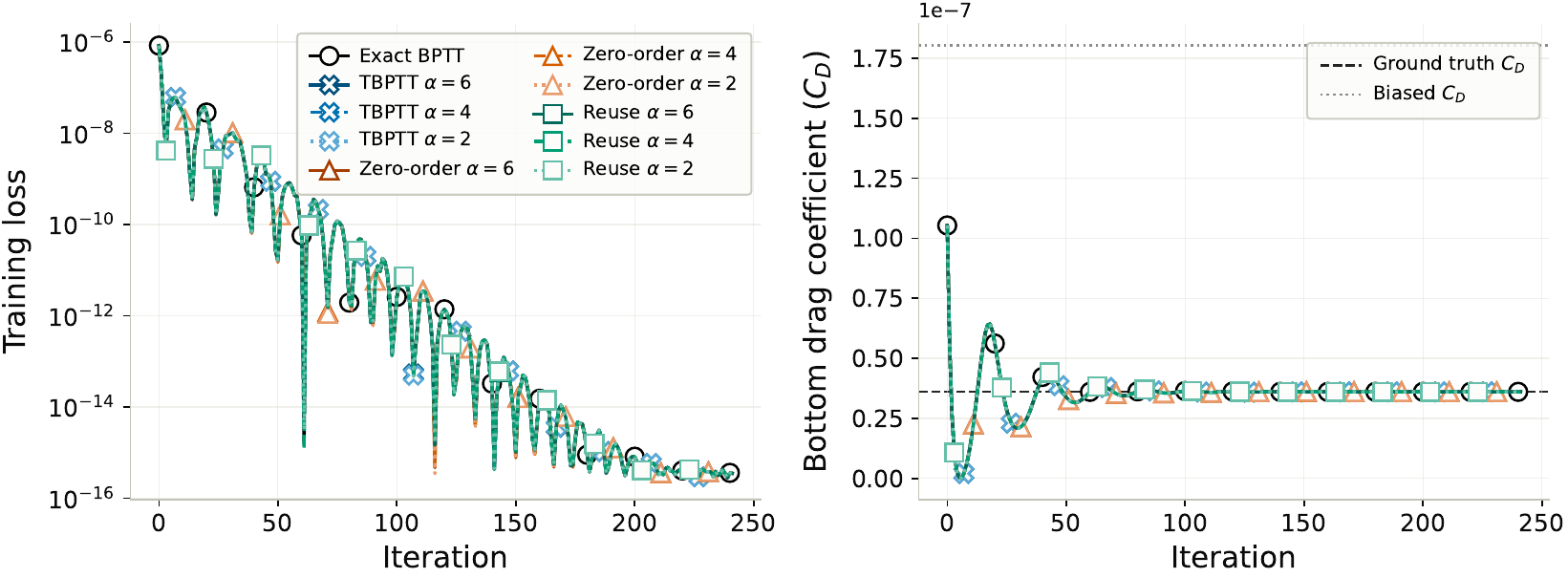}
    \caption{Training loss (left) and evolution of the learned bottom-drag coefficient $C_D$ (right) over 250 optimization iterations, for the exact BPTT, TBPTT, Zero-Order, and Pullback Reuse methods with $\alpha=2$, $4$, $6$. Horizontal lines in the right panel mark the ground truth and biased values of $C_D$. All methods use a rollout of N= 12 steps.}
    \label{fig:train_loss_and_cd_qg}
\end{figure}

Figure~\ref{fig:train_loss_and_cd_qg} shows the training loss and the evolution of the learned bottom drag coefficient $C_D$. All methods move rapidly away from the biased initial value toward the ground truth. The approximate methods fluctuate more from one iteration to the next, with occasional jumps in the loss, but converge to the correct $C_D$ within a comparable number of iterations. This extends the L63 result to a more complex dynamical system, with the optimization problem kept deliberately simple by the use single trainable parameter. The approximate gradients therefore carry enough information to calibrate a physical parameter.

\subsubsection{Neural-network tendency correction}\label{subsubsec:results_ocean_nn}

\begin{figure}[H]
    \centering
    \includegraphics[width=1\linewidth]{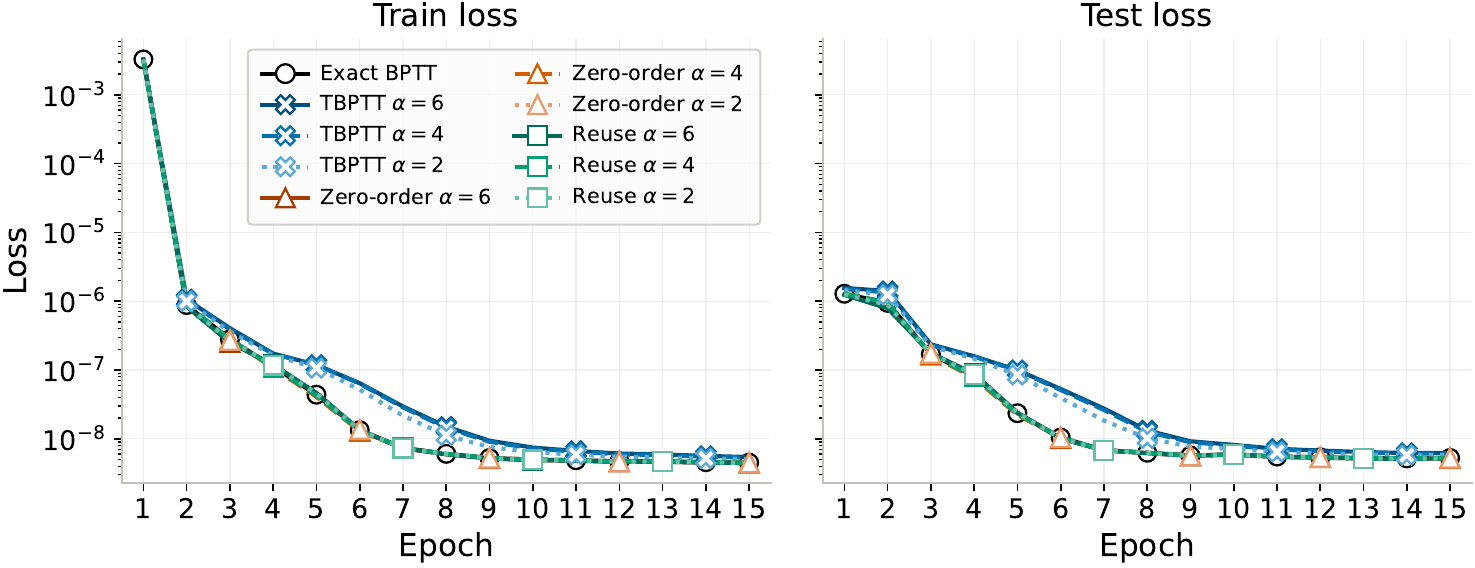}
    \caption{Evolution of the training and test losses over 15 epochs for the exact BPTT, TBPTT, Zero-order, and Reuse Gradient methods with $A=2$, $4$, $6$. For all methods, the CNN was trained with a rollout N=12 steps.}
    \label{fig:train_test_loss_cnn}
\end{figure}

Figure~\ref{fig:train_test_loss_cnn} shows the training and test losses for the CNN correction. All methods converge similarly, with both losses decreasing rapidly toward a common low-loss regime and no systematic degradation as the number of anchor points is reduced. The loss of the CNN trained with TBPTT however, remains above the other methods for part of the optimization before catching up. As in L63, approximate gradients can alter the optimization path without preventing convergence to a comparable final regime. The fact that this also holds for a high-dimensional CNN correction indicates that their usefulness is not limited to single-parameter calibration.

\begin{figure}[H]
    \centering
    \includegraphics[width=0.80\linewidth]{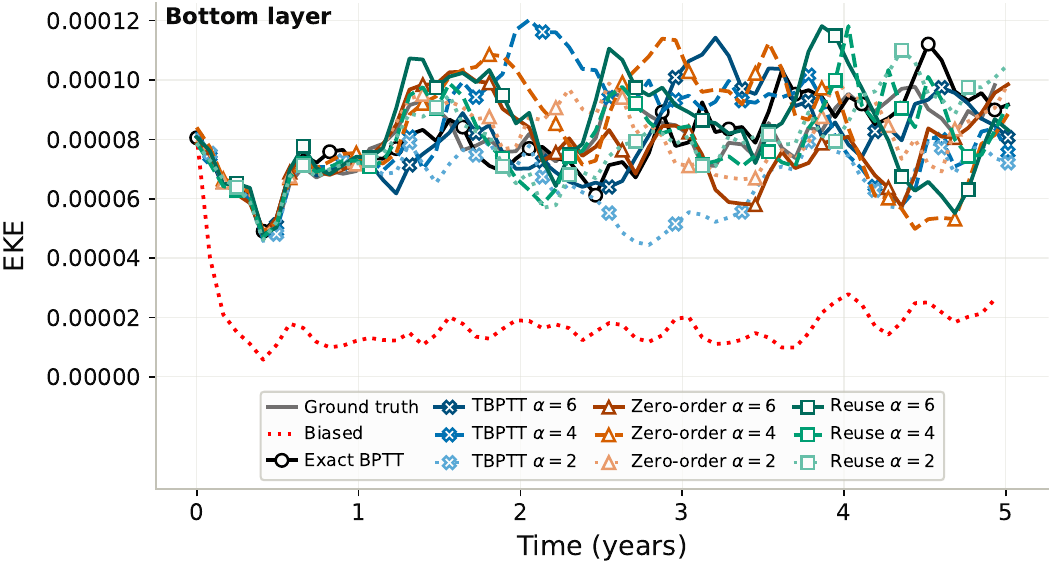}
   \caption{Evolution of the domain-averaged eddy kinetic energy (EKE) in the bottom layer over the 5-year simulation period following training. The CNN was trained with a rollout N=12 steps.}
    \label{fig:evolution_EKE_posttraining}
\end{figure}

Beyond convergence of the online learning objective, we evaluate the trained models in long-term free-running simulations. Figure~\ref{fig:evolution_EKE_posttraining} shows the evolution of the domain-averaged eddy kinetic energy (EKE) in the bottom layer over five years. The biased model exhibits strongly reduced EKE, consistent with its excessive bottom drag, whereas all trained corrections recover values in the same range as the ground truth and hold them for the full five years. The individual trajectories track one another early on before progressively diverging, as expected in a chaotic system.

\begin{figure}[H]
    \centering
    \includegraphics[width=1\linewidth]{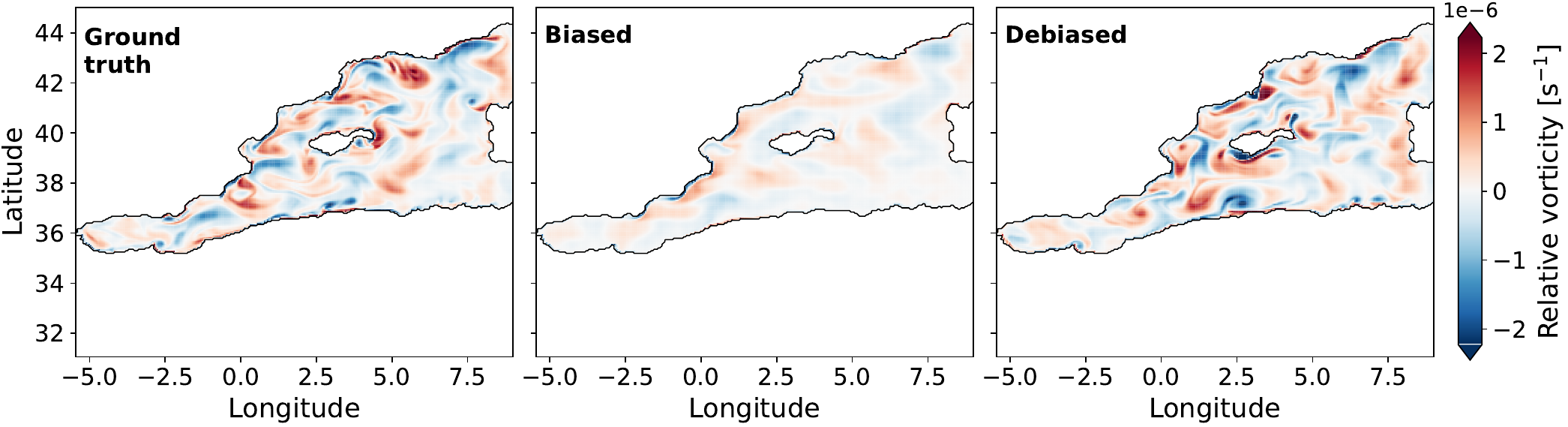}
   \caption{Final bottom-layer relative vorticity after five years of simulation, for the ground-truth, biased and debiased runs. The debiased run applies a CNN correction to the biased model; the CNN was trained over 15 epochs with the Pullback Reuse approximation, a rollout of N = 12 steps and $\alpha$ = 2.}
    \label{fig:final_omega_posttraining}
\end{figure}

To determine whether this recovery extends beyond a bulk statistic, we finally examine the spatial structure of the circulation.  Figure~\ref{fig:final_omega_posttraining} compares the final bottom-layer relative vorticity after five years for the ground-truth, biased and trained simulations. The biased field is markedly weaker, again reflecting the excessive drag, while the trained corrections recover stronger and more coherent vorticity structures, closer to the ground truth.

\section{Discussion and Conclusion}\label{sec:discussion_conclusion}

Differentiable programming is opening new avenues for online learning in numerical modeling, with a growing number of studies using differentiable models to compute sensitivities for model calibration and to train components within a hybrid modeling framework. The computational cost and memory requirements of the backward pass however, grow with both the length of the unrolled trajectory and the dimensionality of the model state, constraining the temporal horizon over which learning can be performed and the spatial resolution at which numerical models that can be trained. 

We propose a class of gradient approximation methods for backpropagation through numerical solvers, that reduce this cost while keeping the gradient accurate enough for optimization. The proposed methods go beyond TBPTT, which discards the sensitivities arriving at truncation boundaries, by preserving the adjoint recurrence over the full trajectory and approximating only the pullback operators within it. Exact pullbacks are constructed only at a sparse set of anchor points. Between anchors, Zero-Order propagates the adjoint through the identity, while Pullback Reuse applies the frozen anchor Jacobian. Since the same parameter pullback is shared across all steps within a chunk, the parameter-gradient sum can be factorized, reducing the number of parameter pullbacks from one per time step to one per chunk.

We evaluated the methods on three systems of increasing complexity, from the low-dimensional Lorenz-63 model to two-dimensional quasi-geostrophic turbulence and a semi-realistic multilayer ocean configuration. In all three, the approximate gradients converge to a loss regime comparable to exact BPTT and recover the reference dynamics, whether the trainable component is a single physical coefficient or a sub-model. Gradient fidelity depends on both the rollout length and the approximation level, with Pullback Reuse consistently giving the highest directional agreement with the exact gradient. On the quasi-geostrophic cases, the proposed methods reduce peak memory by about an order of magnitude, and Zero-order reaches a speedup of roughly four times over exact BPTT while the cosine similarity with the exact gradient stays above 0.7. 

Longer-horizon experiments could reveal an additional effect of pullback approximation beyond computational savings. By changing how sensitivity information propagates through the trajectory, approximate pullbacks may alter the accumulation of unstable gradient components and influence optimization stability. Related work has shown that training-horizon choices and truncated gradient propagation can affect the stability and long-term behavior of learned dynamical systems \citep{list_etal_2022_learned_turbulence_diff}; whether the proposed pullback approximation methods provide a similar stabilizing effect, while retaining more temporal sensitivity information than conventional truncation, remains open. 

The Pullback Approximation is currently controlled by a fixed number of anchors. This gives a simple and reproducible handle on the cost–fidelity trade-off, but applies the same backward resolution throughout the trajectory regardless of the local dynamics. Making the refresh interval adaptive is a natural extension, reconstructing exact pullbacks when an estimate of the approximation error grows too large and reusing them longer when it does not. Possible criteria could include changes in the system's state or energy, measures of local perturbation growth, and indicators of gradient agreement. Such an error-controlled strategy could allocate computational effort where accurate sensitivity information is most important. 

Taken together, these results indicate that substantial reductions in the cost of reverse-mode differentiation are compatible with successful online training of hybrid dynamical models. The proposed methods do not reproduce the exact BPTT gradient, yet the approximate backward passes  retain enough information for optimization while cutting the number of expensive pullback constructions. Zero-Order and Pullback Reuse are two instances of this idea  rather than a closed set, and other backward pass approximations could be explored within the same framework to target different balances between cost and gradient fidelity. If those trade-offs can be exploited effectively, they offer a route toward differentiable online learning at longer rollouts and larger scales than exact BPTT currently permits.

\section*{Acknowledgments}
The authors gratefully acknowledge IMT Atlantique, Lab-STICC, and the Odyssey research team for their support. This work was supported in part by the Région Bretagne and the Brittany doctoral school, and in part by the Horizon Europe project AI4PEX (grant agreement 101137682).

S.O. was supported by the SequoIA AI Cluster Chair HROMS, funded by the Agence Nationale de la Recherche (ANR-23-IACL-0009), and by ANR–France 2030 as part of the PEPR TRACCS program (ANR-25-EXTR-0008). He also received funding from the ANR Chair AI4Extremes and from the French national program LEFE (Les Enveloppes Fluides et l’Environnement) under project OL4ROMS.

\section*{Code and Data Availability}
No observational data were used in this study. All datasets used for training
and evaluation are synthetic and generated by the simulation code. The code
implementing the gradient approximation methods and the numerical experiments
reported in this manuscript is publicly available at:

\url{https://github.com/emiliogz0/approx_adjoint_models_online_learning.git}

% \bibliographystyle{bibstyle}
% \bibliography{references} 

\appendix
\renewcommand{\thesection}{Appendix \Alph{section}:}

\section{Online objectives for the considered use cases}
\label{app:online_objectives}

\paragraph{Lorenz-63 system.}
The online objective is the mean squared error between the state predicted by the
hybrid model, $\vect{u}_{\mathrm{hyb}}(t_n)$, and the reference trajectory generated
by the classical Lorenz-63 system, $\vect{u}_{\mathrm{ref}}(t_n)$, with
$\vect{u} = (x, y, z)^{\top}$. The error is accumulated over the rollout of length
$N$ and averaged over the three state variables,
\begin{equation}
\mathcal{L}
=
\frac{1}{3N}
\sum_{n=1}^{N}
\left\|
\vect{u}_{\mathrm{hyb}}(t_n) - \vect{u}_{\mathrm{ref}}(t_n)
\right\|_2^2 .
\label{eq:loss_l63}
\end{equation}

\paragraph{Large-eddy simulation of quasi-geostrophic turbulence.}
The online objective is the mean squared error between the the filtered vorticity 
predicted by the hybrid LES, $\bar{\omega}_{\mathrm{hyb}}$ and the filtered DNS
reference, $\bar{\omega}_{\mathrm{ref}}$ accumulated over the rollout of length $N$
and averaged over the LES grid $\Omega$,
\begin{equation}
\mathcal{L}
=
\frac{1}{N\,|\Omega|}
\sum_{n=1}^{N}
\sum_{(i,j)\in\Omega}
\Big(
\bar{\omega}_{\mathrm{hyb}}(t_n; i,j) - \bar{\omega}_{\mathrm{ref}}(t_n; i,j)
\Big)^2 ,
\label{eq:loss_qg_les}
\end{equation}
where $|\Omega| = 64 \times 64$ is the number of LES grid points. 

\paragraph{Multilayer quasi-geostrophic Western Mediterranean experiment.}
For both the bottom-drag calibration and the neural-correction experiments, the online objective is the normalized mean squared error between the hybrid-model and reference trajectories. The error is evaluated at the $N_o$ observation times $\tau_k$ within the rollout window and over the full prognostic state, comprising the horizontal velocities
$(u, v)$ and the layer-thickness anomaly $h$,
\begin{equation}
\mathcal{L}
=
\frac{1}{3N_o}
\sum_{k=1}^{N_o}
\sum_{\phi \in \{u,v,h\}}
\frac{1}{L\,|\Omega_\phi|}
\sum_{\ell=1}^{L}
\sum_{(i,j)\in\Omega_\phi}
\left(
\frac{
\phi^\ell_{\mathrm{hyb}}(\tau_k; i,j) - \phi^\ell_{\mathrm{ref}}(\tau_k; i,j)
}{
\sigma_{\phi,\ell}
}
\right)^2 ,
\label{eq:loss_ocean}
\end{equation}
where $\phi^\ell$ denotes field $\phi$ in layer $\ell$, $L$ is the number of layers, and $\Omega_\phi$ is the set of ocean grid points associated with field $\phi$, land points being excluded from the loss. The normalization $\sigma_{\phi,\ell}$ is the
root-mean-square amplitude of field $\phi$ in layer $\ell$, computed over ocean points in the training set.

\section{Simulation configuration and hyperparameters}
\label{app:model_params}

\begin{table}[H]
\centering
\caption{Training setup and hyperparameters for the Lorenz-63 experiment.}
\label{tab:training_setup_l63}
\begin{tabular}{l l}
\toprule
\textbf{Parameter} & \textbf{Value} \\
\midrule
Numerical solver & RK4 \\
Loss function    & MSE \\
Optimizer        & Adam \\
Learning rate    & $10^{-3}$ (constant) \\
Rollout length   & 12 \\
Batch size       & 20 \\
Epochs           & 250 \\
\bottomrule
\end{tabular}
\end{table}

\begin{table}[H]
\centering
\caption{Training setup and hyperparameters for the quasi-geostrophic turbulence
large-eddy simulation experiment.}
\label{tab:training_setup_simple_gq}
\begin{tabular}{l l}
\toprule
\textbf{Parameter} & \textbf{Value} \\
\midrule
Numerical solver & RK4 \\
Loss function    & MSE \\
Optimizer        & Adam \\
Learning rate    & $10^{-4}$ (constant) \\
Rollout length   & 40 \\
Batch size       & 20 \\
Epochs           & 20 \\
\bottomrule
\end{tabular}
\end{table}

\begin{table}[H]
\centering
\caption{Physical and numerical parameters for the Western Mediterranean
configuration of the multilayer quasi-geostrophic model.}
\label{tab:ocean_reg_config}
\begin{tabular}{l l l}
\toprule
\textbf{Parameter} & \textbf{Value} & \textbf{Description} \\
\midrule
$L_x \times L_y$            & $\approx 1550 \times 1270\,\mathrm{km}$ & Domain size \\
$n_x \times n_y$            & $310 \times 254$ & Grid dimensions \\
$\Delta x \times \Delta y$  & $5 \times 5\,\mathrm{km}$ & Spatial resolution \\
$n_\ell$                    & 3 & Number of vertical layers \\
$H_\ell$                    & $[75,\,400,\,2525]\,\mathrm{m}$ & Reference layer thickness \\
$g'_\ell$                   & $[9.81,\,0.025,\,0.0125]\,\mathrm{m\,s^{-2}}$ & Reduced gravity \\
$C_D$                       & $3.6057 \times 10^{-8}\,\mathrm{s^{-1}}$ & Bottom drag coefficient \\
$\tau_0$                    & $1.3 \times 10^{-3}\,\mathrm{N\,m^{-2}}$ & Wind stress magnitude \\
$\rho_0$                    & $1000\,\mathrm{kg\,m^{-3}}$ & Reference density \\
$f_0$                       & $9.375 \times 10^{-5}\,\mathrm{s^{-1}}$ & Mean Coriolis parameter \\
$\beta$                     & $1.754 \times 10^{-11}\,\mathrm{m^{-1}\,s^{-1}}$ & Coriolis gradient \\
---                         & 0.6 & Slip coefficient (lateral boundaries) \\
$h$                         & $3600\,\mathrm{s}$ & Time step \\
---                         & RK3-SSP & Time-integration scheme \\
\bottomrule
\end{tabular}
\end{table}

\begin{table}[H]
\centering
\caption{Training setup and hyperparameters for the two experiments on the Western
Mediterranean configuration of the multilayer quasi-geostrophic model.}
\label{tab:training_setup}
\begin{tabular}{l c c}
\toprule
\textbf{Parameter} & \textbf{$C_D$ calibration} & \textbf{Sub-model training} \\
\midrule
Numerical solver        & \multicolumn{2}{c}{RK3-SSP} \\
Loss function           & \multicolumn{2}{c}{Normalized MSE, Eq.~\eqref{eq:loss_ocean}} \\
Optimizer               & \multicolumn{2}{c}{Adam} \\
Rollout length $N$      & \multicolumn{2}{c}{12} \\
\addlinespace
Trainable parameters    & $C_D$ (1 scalar) & CNN weights (1456) \\
Learning rate           & $10^{-4}$ (constant) & $10^{-6}$ (cosine schedule) \\
Batch size  &
\multicolumn{2}{c}{1}   \\
Optimization budget     & 250 iterations   & 15 epochs \\
\bottomrule
\end{tabular}
\end{table}

\end{document}

\typeout{get arXiv to do 4 passes: Label(s) may have changed. Rerun}